\documentclass{article} 
\usepackage{iclr2025_conference,times}
\iclrfinalcopy

\usepackage{amsmath,amsfonts,bm}

\def\figref#1{fig.~\ref{#1}}
\def\eqref#1{eq.~\ref{#1}}
\def\1{\bm{1}}

\DeclareMathAlphabet{\mathsfit}{\encodingdefault}{\sfdefault}{m}{sl}
\SetMathAlphabet{\mathsfit}{bold}{\encodingdefault}{\sfdefault}{bx}{n}

\usepackage{hyperref}
\usepackage{url}
\usepackage{microtype}      

\usepackage{graphicx}
\usepackage{booktabs} 
\usepackage{caption}
\usepackage{subcaption}

\usepackage{svg}
\usepackage{wrapfig}
\usepackage{xcolor}
\usepackage{enumitem}

\usepackage{algorithm}
\usepackage{algpseudocode}
\algtext*{EndWhile}
\algtext*{EndIf}

\newcommand{\norm}[1]{\left| \left| #1 \right| \right|}

\definecolor{ntcol}{rgb}{0.0, 0.62, 0.24}

\newcommand{\nt}[1]{{\color{black}#1}}

\newcommand{\bh}[1]{{\color{black}#1}}

\newcommand{\paraspace}{\vspace{-3pt}}

\title{Flow Matching for Posterior Inference with Simulator Feedback}

\author{Benjamin Holzschuh, Nils Thuerey %\thanks{ Use footnote for providing further information about author (webpage, alternative address)---\emph{not} for acknowledgingfunding agencies.  Funding acknowledgements go at the end of the paper.} 
\\
School for Computation, Information and Technology\\
Technical University of Munich \\
\texttt{\{benjamin.holzschuh, nils.thuerey\}@tum.de} \\
}

\newcommand{\myskip}{\vspace{-5pt}} 

\begin{document}

\maketitle

\begin{abstract}
Flow-based generative modeling is a powerful tool for solving inverse problems in physical sciences that can be used for sampling and likelihood evaluation with much lower inference times than traditional methods. 
We propose to refine flows with additional control signals based on a simulator. Control signals can include gradients and a problem-specific cost function if the simulator is differentiable, or they can be fully learned from the simulator output.
In our proposed method, we pretrain the flow network and include feedback from the simulator exclusively for finetuning, therefore requiring only a small amount of additional parameters and compute. We motivate our design choices on several benchmark problems for simulation-based inference and evaluate flow matching with simulator feedback against classical MCMC methods for modeling strong gravitational lens systems, a challenging inverse problem in astronomy. We demonstrate that including feedback from the simulator improves the accuracy by $53\%$, making it competitive with traditional techniques while being up to 67x faster for inference. Code and experiments are available at \url{https://github.com/tum-pbs/sbi-sim}.
\end{abstract}

\paraspace{}
\section{Introduction}

Acquiring posterior distributions given measurement data is of paramount scientific interest \citep{cranmer2020frontier}, with real-world applications ranging from particle physics \citep{DBLP:conf/nips/BaydinSBHNM0GLM19}, over the inference of gravitational waves \citep{dax2021real} to predictions of dynamical systems such as weather forecasting \citep{gneiting2005weather}. 
For an observation $\boldsymbol{x}_o$ and model parameters $\boldsymbol{\theta}$, the likelihood $p(\boldsymbol{x}_o|\boldsymbol{\theta})$ corresponds to how strongly we believe a model with parameters $\boldsymbol{\theta}$ causes $\boldsymbol{x}_o$ to occur. In Bayesian modeling, we are interested in the posterior $p(\boldsymbol{\theta}|\boldsymbol{x}_o)$, which is proportional to the likelihood times the prior $p(\boldsymbol{\theta})$ and tells us which parameters most likely explain the observation. 

Inferring the posterior based on samples from traditional likelihood-based methods can be expensive for high-dimensional data and when likelihood evaluations are costly. Simulation-based inference \citep[SBI]{cranmer2020frontier} represents the posterior as a parametric function $q(\boldsymbol{\theta}|\boldsymbol{x}_o)$, which is a learnable conditional density estimator that can be trained purely by simulations $\boldsymbol{x} \sim p(\boldsymbol{x}|\boldsymbol{\theta})$ alone. 
By investing an upfront cost for training the density estimator, we can sample and compute likelihoods from $q(\boldsymbol{\theta}|\boldsymbol{x}_o)$ much faster than other methods, thereby amortizing the training cost over many observations.

Traditionally, normalizing flows \citep{DBLP:conf/icml/RezendeM15, DBLP:conf/iclr/DinhSB17, DBLP:conf/aistats/PapamakariosS019} have been a popular class of density estimators used in many areas of science. To compute likelihoods and for sampling, normalizing flows transform a noise distribution to the posterior distribution. \bh{With the success of diffusion models \cite{DBLP:conf/nips/HoJA20, DBLP:conf/iclr/LiuG023, DBLP:conf/iclr/LipmanCBNL23}, it became clear} that the mapping between sampling and posterior distribution can be specified a priori, for example by specifying a corruption process that transforms any data distribution to a normal Gaussian. The resulting continuous-time models outperform discrete architecture in many areas, and training larger models is much more scalable. 

\begin{figure}[t]
    \centering
    \includegraphics[width=\textwidth]{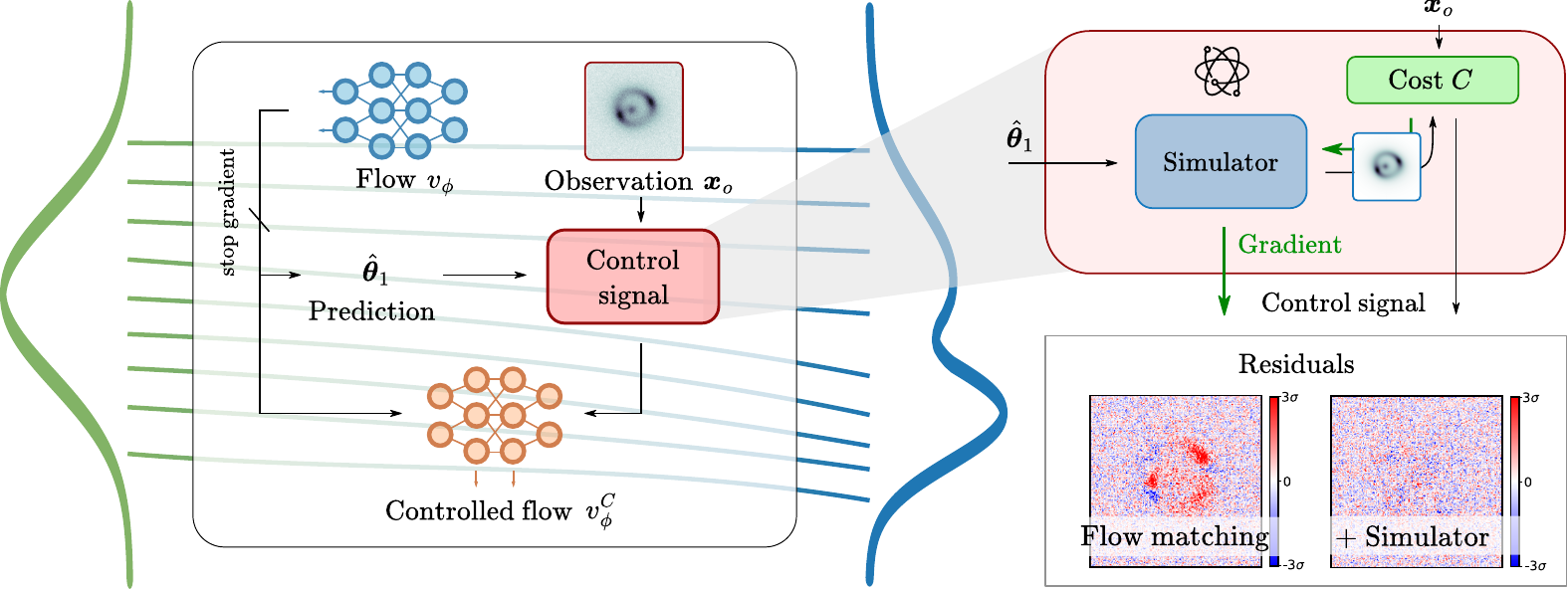}
    \caption{An overview of our proposed framework. We consider a pretrained flow network $v_\phi$ and use the predicted flow for the trajectory point $\boldsymbol{\theta}_t$ at time $t$ to estimate \smash{$\hat{\boldsymbol{\theta}}_1$}.  
    \bh{On the right, we show a gradient-based control signal with a differentiable simulator and cost function $C$ for improving \smash{$\hat{\boldsymbol{\theta}}_1$}}.
    An additional network learns to combine the predicted flow with feedback via the control signal to give a new controlled flow.
    By combining learning-based updates with suitable controls, we avoid local optima and obtain high-accuracy samples with low inference times.}\vspace{-0.21cm}
    \label{fig:hybrid_model}
\end{figure}

Despite the widespread success of flow-based models for generative modeling and density estimation, there is no direct feedback loop between the model, the observation $\boldsymbol{x}_o$ and the sample $\boldsymbol{\theta}$ during training, which makes it very difficult to produce highly accurate samples based on learning alone.

We propose a simple strategy to reintroduce control signals using \textbf{simulators} into the flow network. We refine an existing pretrained flow-based model with a flexible 
control signal by aggregating the learned flow and control signals into a \textit{controlled flow}. The aggregation network is small compared to the pretrained flow network, and we find that freezing the weights of the pretrained network works very well; thus, refining needs only a minimal amount of additional parameters and compute.

To demonstrate how these refinements affect the accuracy of samples and the posterior, we consider modeling strong gravitational lens systems
\citep{hezaveh2017fast,cunha2018shadows,legin2021simulation, vegetti2009bayesian, vegetti2023strong}, 
an inverse problem in astrophysics that is challenging and requires precise posteriors for accurate modeling of observations. In galaxy-scale strong lenses, light from a source galaxy is deflected by the gravitational potential of a galaxy between the source and observer, causing multiple images of the source to be seen. Since these images and their distortions are sensitive to the distribution of matter on small scales, this can act as a probe for different dark matter models. 
With upcoming and current sky surveys \citep{laureijs2011euclid} expected to release large data catalogs in the near future, the number of known lenses will increase dramatically by several orders of magnitude. Traditional computational approaches require several minutes to many hours or days to model a single lens system. Therefore, there is an urgent need to reduce the compute and inference with learning-based methods. In this experiment, we demonstrate that using flow matching and our proposed control signals with feedback from a simulator, we obtain posterior distributions for lens modeling that are competitive with the posteriors obtained by MCMC-based methods but with much faster inference times. 

Additionally, we evaluate different related variants of flow matching such as independent couplings \citep{DBLP:journals/corr/tong2023}, self-conditioning \citep{DBLP:conf/iclr/ChenZH23} or different loss formulations in the context of SBI using several benchmark problems. We then analyze our proposed control signals for the Lotka-Volterra model, a system of coupled ordinary differential equations (ODEs) descriping the population evolution of predators and prey over time. Our analysis underscores the essential role of simulator feedback for inference and that high accuracy is very challenging to achieve from scaling up datasets and model sizes alone.

To summarize, the main contributions of our work are:

\vspace{-2pt}
\begin{itemize}
\setlength\itemsep{-1pt} 
\item We propose a versatile strategy to improve pretrained flows with control signals based on feedback from a simulator. Control signals can be based on gradients and a cost function, if the simulator is differentiable, but they can also be learned directly from the simulator output.

\item We assess different variants of flow matching in the context of SBI and demonstrate with the Lotka-Volterra model that performance gains due to simulator feedback are substantial and cannot be achieved by training on larger datasets alone.

\item We demonstrate the efficacy of our proposed finetuning with control signals for inferring the parameter distributions of strong gravitational lens systems, a challenging inverse problem in astronomy that is sensitive to sample accuracy. We show that flow matching with simulator feedback is competitive with MCMC baselines and beats them significantly regarding inference time.
\end{itemize}

\section{Related Work}

\paraspace{}\paragraph{Solving inverse problems with diffusion models}
Diffusion models have been proposed to solve linear inverse problems \citep{DBLP:conf/nips/KawarVE21, DBLP:conf/nips/KawarEES22, DBLP:conf/cvpr/ChungSY22, DBLP:journals/corr/cardoso2023}, as well as general inverse problems \citep{DBLP:conf/nips/HolzschuhVT23, DBLP:conf/iclr/ChungKMKY23, DBLP:conf/nips/ChungKY23}. 
\bh{In most of these works, the diffusion model learns the prior distribution and sampling from the posterior is achieved through a modified inference procedure, which guides samples via a conditioning}. The conditioning can be based on a class label, text input \citep{DBLP:conf/iclr/songscore21, DBLP:journals/corr/hoclassifierfree2022, DBLP:conf/nips/SahariaCSLWDGLA22, DBLP:conf/nips/WuTNBC23} or directly on a differentiable measurement operator \citep{DBLP:conf/iclr/ChungKMKY23, DBLP:conf/nips/ChungKY23}. 
In contrast to these works, we finetune a pretrained flow and learn an optimal combination of the pretrained flow and feedback from a simulator via control signals in the broader flow matching context.

\paraspace{}\paragraph{Flow matching}
Our work builds on top of prior work in flow matching \citep{DBLP:conf/iclr/LipmanCBNL23, DBLP:journals/corr/albergo2023, DBLP:conf/icml/PooladianBDALC23, DBLP:journals/corr/tong2023, DBLP:journals/corr/albergo2023coupling}, particularly we adopt and evaluate conditional optimal transport paths \citep{DBLP:conf/iclr/LipmanCBNL23} and independent couplings or rectified flows \citep{DBLP:conf/iclr/LiuG023, DBLP:journals/corr/tong2023} for simulation-based inference. We extend the existing literature by adding feedback from a simulator for scientific inverse problems. 

\paraspace{}\paragraph{Simulation-based inference} Our work directly compares to neural posterior estimation approaches for simulation-based inference \citep[SBI]{cranmer2020frontier, DBLP:conf/aistats/LueckmannBGGM21}. Contrary to static architectures \citep{DBLP:conf/iclr/DinhSB17, DBLP:conf/nips/KingmaD18, DBLP:conf/nips/PapamakariosMP17, DBLP:conf/nips/DurkanB0P19}, our approach extends the continuous-time paradigm \citep{DBLP:conf/nips/ChenRBD18, DBLP:conf/iclr/GrathwohlCBSD19}. \cite{DBLP:conf/nips/WildbergerDBGMS23} have applied flow matching to neural posterior estimation and \cite{DBLP:journals/corr/sharrock2022} have used conditional diffusion models and Langevin dynamics during sampling. In contrast to previous work, we include controls signals via problem-specific simulators and cost functions during training to significantly improve the sampling quality.

\paraspace{}\paragraph{Strong lensing and parameter estimation} Machine learning has been successfully applied to estimate parameters of lens and source models \citep{hezaveh2017fast, levasseur2017uncertainties}, however, previous methods are usually restricted to point estimates, use simple variational distributions or Bayesian Neural Networks \citep{schuldt2021holismokes, legin2021simulation, legin2023framework, poh2022strong} that are not well suited to represent more complicated high-dimensional data distributions. In this paper, we combine flow matching with problem-specific simulators to obtain highly accurate samples via feedback from control signals.

\section{Flow Matching Theory}

Continuous-time flow models transform samples $\boldsymbol{\theta}$ from a sampling distribution $p_0$ to samples of a target or posterior distribution $p_1$. This mapping can be expressed via the ODE
\begin{align} \label{eq:neural_ode}
    d\boldsymbol{\theta}_t = v_\theta(t,\boldsymbol{\theta}_t) dt,
\end{align} 
where $v_\phi(t,\boldsymbol{\theta}_t)$ represents a neural network with parameters $\phi$. Early works \citep{DBLP:conf/nips/ChenRBD18, DBLP:conf/iclr/GrathwohlCBSD19} optimize $v_\phi(t,\boldsymbol{\theta})$ using maximum likelihood training, which is computationally demanding and difficult to scale to larger networks. 
Instead, in flow matching the network $v_\phi(t,\boldsymbol{\theta})$ is trained by regressing a vector field $u(t,\boldsymbol{\theta})$ that generates probability paths that map from $p_0$ to $p_1$. 

\paraspace{}\paragraph{Generating probability paths}
We say that a smooth vector field $u : [0,1] \times \mathbb{R}^d \to \mathbb{R}^d$, called \emph{velocity}, generates the probability paths $p_t$, if it satisfies the continuity equation $\frac{\partial p}{\partial t} = - \nabla \cdot (p_t u_t)$. 
Informally, this means that we can sample from the distribution $p_t$ by sampling $\boldsymbol{\theta}_0 \sim p_0$ and then solving the ODE $d\boldsymbol{\theta} = u(t,\boldsymbol{\theta})dt$ with initial condition $\boldsymbol{\theta}_0$. In the following, we will denote $u(t,\boldsymbol{\theta})$ by $u_t(\boldsymbol{\theta})$.
To regress the velocity field, we define the \textbf{flow matching} objective
\begin{align} \label{eq:flow_matching}
    \mathcal{L}_\mathrm{FM}(\theta) := \mathbb{E}_{t\sim \mathcal{U}(0,1), \boldsymbol{\theta} \sim p_t(\boldsymbol{\theta})} \norm{v_\theta(t,\boldsymbol{\theta}) - u_t(\boldsymbol{\theta})}^2.
\end{align}
In order to compute this loss, we need to sample from the probability distribution $p_t(\boldsymbol{\theta})$ and we need to know the velocity $u_t(\boldsymbol{\theta})$. However, in general $u_t(\boldsymbol{\theta})$ is not accessible.

\paraspace{}\paragraph{Conditioning variable}
To solve this problem, we apply a trick by introducing a latent variable $\boldsymbol{z}$ distributed according to $q(\boldsymbol{z})$ and define the conditional likelihoods $p_t(\boldsymbol{\theta}|\boldsymbol{z})$ that depend on the latent variable so that $p_t(\boldsymbol{\theta}) = \int p_t(\boldsymbol{\theta}|\boldsymbol{z})q(\boldsymbol{z})d\boldsymbol{z}$. Interestingly, if the conditional likelihoods are generated by the velocities $u_t(\boldsymbol{\theta}|\boldsymbol{z})$, then the velocity $u_t(\boldsymbol{\theta})$ can be written in terms of $u_t(\boldsymbol{\theta}|\boldsymbol{z})$ and $p_t(\boldsymbol{\theta}|\boldsymbol{z})$ with $u_t(\boldsymbol{\theta}) := \mathbb{E}_{q(\boldsymbol{z})}[u_t(\boldsymbol{\theta}|\boldsymbol{z})p_t(\boldsymbol{\theta}|\boldsymbol{z})/p_t(\boldsymbol{\theta})]$. We can choose paths $p_t(\boldsymbol{\theta}|\boldsymbol{z})$ that are easy to sample from and for which we know the generating velocities $u_t(\boldsymbol{\theta}|\boldsymbol{z})$.   
Next, we define the \textbf{conditional flow matching} loss
\begin{align} \label{eq:conditional_flow_matching}
    \mathcal{L}_\mathrm{CFM}(\phi) := \mathbb{E}_{t,q(\boldsymbol{z}),p_t(\boldsymbol{\theta}|\boldsymbol{z})} \norm{v_\phi(t,\boldsymbol{\theta})-u_t(\boldsymbol{\theta}|\boldsymbol{z})}^2.
\end{align}
In contrast to the flow matching loss \eqref{eq:flow_matching}, this loss is tractable and can be used for optimization. Now, one can show \citep{DBLP:journals/corr/tong2023} that if $p_t(\boldsymbol{\theta}) > 0$ for all $\boldsymbol{\theta} \in \mathbb{R}^d$, then
\begin{align}
    \nabla_\phi \mathcal{L}_\mathrm{FM}(\phi) = \nabla_\phi \mathcal{L}_\mathrm{CFM}(\phi).
\end{align}
This means that we can train $v_\theta(\boldsymbol{\theta},t)$ to regress $u_t(\boldsymbol{\theta})$ generating the mapping between $p_0$ and $p_1$ by optimizing the conditional flow matching loss \eqref{eq:conditional_flow_matching}.

\paraspace{}\paragraph{Couplings} 
The above framework allows for many degrees of freedom when specifying the mapping from $p_0$ to $p_1$ via the conditioning variable $\boldsymbol{z}$ and the conditional likelihoods $p_t$. 
One particularly intuitive and simple choice is to consider the coupling $q(\boldsymbol{z}) = p_1(\boldsymbol{\theta})$  \citep{DBLP:conf/iclr/LipmanCBNL23} together with conditional probability and generating velocity \begin{align} \label{eq:conditional_ot_prob}
    p_t(\boldsymbol{\theta}|\boldsymbol{\theta}_1) &= \mathcal{N}(\boldsymbol{\theta} |\, t \boldsymbol{\theta}_1, (1-(1-\sigma_\mathrm{min})t)I) \\ \label{eq:conditional_ot_velocity}
    u_t(\boldsymbol{\theta}|\boldsymbol{\theta}_1) &= \frac{\boldsymbol{\theta}_1-(1-\sigma_\mathrm{min})\boldsymbol{\theta}}{1-(1-\sigma_\mathrm{min})t},
\end{align}
where $\sigma_\mathrm{min} > 0$. Conditioned on $\boldsymbol{\theta}_1$, this coupling transports a point $\boldsymbol{\theta}_0 \sim \mathcal{N}(0,I)$ from the sampling distribution to the posterior distribution on the linear trajectory $t \boldsymbol{\theta}_1$ ending in $\boldsymbol{\theta}_1$ but decreasing the standard deviation from $1$ to a smoothing constant $\sigma_\mathrm{min}$. In this case, the transport path coincides with the optimal transport between two Gaussian distributions. 

\section{Controls for Improved Accuracy} \label{sec:controls}
While flow-based models $v_\phi(t,\boldsymbol{\theta})$ gradually transform samples from $p_0$ to $p_1$ in many steps during inference via solving the ODE \eqref{eq:neural_ode}, there is no direct feedback loop between the underlying simulator, the current point on the trajectory $\boldsymbol{\theta}_t$, and the observation $\boldsymbol{x}_o$. 
A central goal of our work is to reintroduce this feedback loop into inference and training by incorporating a control signal. 

\paraspace{}\paragraph{Conditioning of flows} Flows $v_\phi(t,\boldsymbol{\theta})$ can be conditioned on an observation $\boldsymbol{x}_o$ through an additional input $v_\phi(t,\boldsymbol{\theta},\boldsymbol{x}_o)$, therefore modeling the conditional densities $p_t(\boldsymbol{\theta}|\boldsymbol{x}_o)$ \citep{DBLP:conf/iclr/songscore21}. For example, in classifier free-guidance \citep{DBLP:journals/corr/hoclassifierfree2022}, this conditioning is randomly dropped and set to $0$ during training. The resulting models can then be used for both conditional and unconditional generation.   

A fundamental problem here is that the conditioning $\boldsymbol{x}_o$ is static, whereas 
we propose to have a dynamic control mechanism that depends on the trajectory $\boldsymbol{\theta}_t$, the observation, and an underlying control signal. The latter should relate $\boldsymbol{\theta}_t$ and observation using a physics-based model represented through a cost function $C$. As the accuracy of neural networks is inherently limited by the finite size of their weights, and smaller networks are attractive from a computational perspective, physics-based control has the potential to yield high accuracy with lean and efficient neural network models.

\paraspace{}\paragraph{1-step prediction} 
An additional issue is that the current trajectory $\boldsymbol{\theta}_t$ might not be close to a good estimate of a posterior sample $\boldsymbol{\theta}_1$, especially at the beginning of inference, where $\boldsymbol{\theta}_0$ is drawn from the sampling distribution.
This issue is alleviated by applying the cost function $C$ to the current estimate $\boldsymbol{\theta}_t$, we extrapolate $\boldsymbol{\theta}_t$ forward in time to obtain an estimated \smash{$\hat{\boldsymbol{\theta}}_1$} 
\begin{align} \label{eq:1_step_prediction}
    \hat{\boldsymbol{\theta}}_1 = \boldsymbol{\theta}_t + (1-t) v_\phi(t, \boldsymbol{\theta}_t, \boldsymbol{x}_o).
\end{align}

\paraspace{}\paragraph{Comparison with likelihood-guidance} The 1-step prediction is conceptually related to diffusion sampling using likelihood-guidance \citep{DBLP:conf/cvpr/ChungSY22, DBLP:conf/nips/WuTNBC23}. For inference in diffusion models, sampling is based on the conditional score $\nabla_{\boldsymbol{\theta}_t} \log p(\boldsymbol{\theta}_t|\boldsymbol{x}_o)$, which can be decomposed into \begin{align}
    \nabla_{\boldsymbol{\theta}_t} \log p(\boldsymbol{\theta}_t|\boldsymbol{x}_o) = \nabla_{\boldsymbol{\theta}_t} \log p(\boldsymbol{\theta}_t) + \nabla_{\boldsymbol{\theta}_t} \log p(\boldsymbol{x}_o|\boldsymbol{\theta}_t).
\end{align}
The first expression can be estimated using a pretrained diffusion model, whereas the latter is usually intractable, but can be approximated using 
$p(\boldsymbol{x}_o|\boldsymbol{\theta}_t) \approx p_{\boldsymbol{x}_o|\boldsymbol{\theta}_0}(\boldsymbol{x}_o|\hat{\boldsymbol{\theta}}(\boldsymbol{\theta}_t))$,
where the denoising estimate \smash{$\hat{\boldsymbol{\theta}}(\boldsymbol{\theta}_t) := \mathbb{E}_q[\boldsymbol{\theta}_0|\boldsymbol{\theta}_t]$} is usually obtained via Tweedie's formula $(\mathbb{E}_q[\boldsymbol{\theta}_0|\boldsymbol{\theta}_t] - \boldsymbol{\theta}_t) / t\sigma^2$. In practice, the estimate \smash{$\hat{\boldsymbol{\theta}}(\boldsymbol{\theta}_t)$} is very poor when $\boldsymbol{\theta}_t$ is still noisy, impeding the inference in the early stages. On the contrary, flows based on linear conditional transportation paths have empirically been shown to have trajectories with less curvature \citep{DBLP:conf/iclr/LipmanCBNL23} compared to, for example, diffusion models, thus enabling inference in fewer steps and providing better estimates for \smash{$\hat{\boldsymbol{\theta}}_1$}.

\paraspace{}\paragraph{Controlled flow $v_\phi^C$} We pretrain the flow network $v_\phi(t,\boldsymbol{\theta},\boldsymbol{x}_o)$ without any control signals to make sure that we can realize the best achievable performance possible based on learning alone. Then, in a second training phase, we introduce the control network $v_\phi^C(t, \boldsymbol{v}, \boldsymbol{c})$ with pretrained flow $\boldsymbol{v}$ and control signal $\boldsymbol{c}$ as input. The control network is much smaller in size than the flow network, making up ca. $10\%$ of the weights $\phi$ in our large-scale experiments. We freeze the network weights of $v_\phi$ and train with the conditional flow matching loss \eqref{eq:conditional_flow_matching} for a small number of additional steps. This reduces training time and compute since we do not need to backpropagate gradients through $v_\phi(t, \boldsymbol{\theta}, \boldsymbol{x}_o)$. We did not observe that freezing the weights of $v_\phi$ affects the performance negatively. We include algorithms for training in appendix \ref{app:algorithms}.

\subsection{Types of control signals}

\begin{wrapfigure}{r}{55mm}
    \begin{minipage}{\linewidth}
    \centering
    \vspace{-.5cm}
    \includegraphics[width=.92\textwidth]{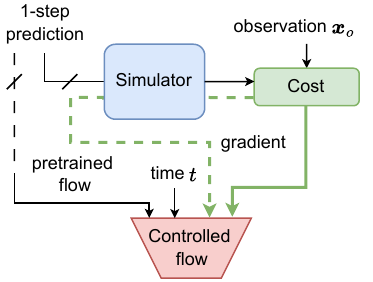}
    \subcaption{Gradient-based control signal}
    \label{fig:gradient_based_control_signal}
    \vspace{.2cm}
    \includegraphics[width=.92\textwidth]{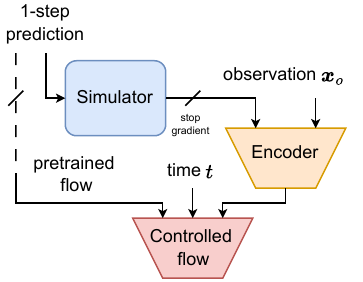}
    \subcaption{Learning-based control signal}
    \label{fig:learning_based_control_signal}
    \end{minipage}
    \caption{Control signals with simulator feedback.}
    \label{fig:types_of_control_signals}
\end{wrapfigure}

Aiming for high inference accuracy, we extend self-conditioning via physics-based control signals to include an additional feedback loop between the model output and an underlying physics-based prior. We distinguish between two types of control signals. 

\paraspace{}\paragraph{Gradient-based control signal} 

In the first case, there is a differentiable cost function $C$ and a deterministic differentiable simulator $S$ as shown in \figref{fig:gradient_based_control_signal}.
Given an observation $\boldsymbol{x}_o$ and the estimated prediction \smash{$\hat{\boldsymbol{\theta}}_1$}, the control signal relates to how well \smash{$\hat{\boldsymbol{\theta}}_1$} explains $\boldsymbol{x}_o$ via some cost function $C$. The cost function can also depend directly on or be equal to the likelihood \smash{$p(\boldsymbol{x}_o|\hat{\boldsymbol{\theta}}_1)$}. For a differentiable cost function $C$, we define the control signal via  
    \begin{align}
    \boldsymbol{c}(\hat{\boldsymbol{\theta}}_1, \boldsymbol{x}_o) := [C(S(\hat{\boldsymbol{\theta}}_1), \boldsymbol{x}_o); \nabla_{\hat{\boldsymbol{\theta}}_1} C(S(\hat{\boldsymbol{\theta}}_1), \boldsymbol{x}_o)].
    \end{align}
Note that while we recommend using informative control signals, we can use any control that depends on \smash{$\hat{\boldsymbol{\theta}}_1$} and $\boldsymbol{x}_o$. 

\paraspace{}\paragraph{Learning-based control signal}

In the second case, the simulator is non-differentiable. To combine the simulator output with the observation $\boldsymbol{x}_o$, we introduce a learnable encoder model \textit{Enc} with parameters $\phi_E$. The output of the encoder is small and of size $O(\mathrm{dim}(\boldsymbol{\theta}))$.
The control signal is then defined as 
\begin{align}
    \boldsymbol{c}(\hat{\boldsymbol{\theta}}_1, \boldsymbol{x}_o) := Enc(S(\hat{\boldsymbol{\theta}}_1), \boldsymbol{x}_o).
\end{align}
The gradient backpropagation is stopped at the simulator output, see \figref{fig:learning_based_control_signal}. 

\subsection{Additional Considerations for Simulator Feedback}

\paraspace{}\paragraph{Stochastic simulators}

Many Bayesian inference problems have a stochastic simulator. For simplicity, we assume that all stochasticity within such a simulator can be controlled via a variable $z \sim \mathcal{N}(0, I)$, which is an additional input. Motivated by the equivalence of exchanging expectation and gradient  
\begin{align}
    \nabla_{\hat{\boldsymbol{\theta}}_1} \mathbb{E}_{z\sim \mathcal{N}(0,1)} [ C(S_z(\hat{\boldsymbol{\theta}}_1), \boldsymbol{x}_o)] = \mathbb{E}_{z\sim \mathcal{N}(0,1)} [ \nabla_{\hat{\boldsymbol{\theta}}_1} C(S_z(\hat{\boldsymbol{\theta}}_1), \boldsymbol{x}_o)],
\end{align}
when calling the simulator, we draw a random realization of $z$. During training, we randomly draw $z$ for each sample and step while during inference we keep the value of $z$ fixed for each trajectory. 

\paraspace{}\paragraph{Time-dependence}
If the estimate $\hat{\boldsymbol{\theta}}_1$ is bad and the corresponding cost $C(\hat{\boldsymbol{\theta}}_1, \boldsymbol{x}_o)$ is high, gradients and control signals can become unreliable. In appendix \ref{sec:sbi}, we empirically find that the estimates \smash{$\hat{\boldsymbol{\theta}}_1$} become more reliable for $t \geq 0.8$. Therefore, we only train the control network \smash{$v_\phi^C$} in this range, which allows for focusing on control signals containing the most useful information. For $t < 0.8$, we directly output the pretrained flow $v_\phi(t, \boldsymbol{\theta}, \boldsymbol{x}_o)$.

\paraspace{}\paragraph{Theoretical correctness} Contrary to likelihood-based guidance, which uses an approximation for \smash{$\nabla_{\boldsymbol{\theta}_t} \log p(\boldsymbol{x}_o|\boldsymbol{\theta}_t)$} as a guidance term during inference, the approximation \smash{$\hat{\boldsymbol{\theta}}_1$} only influences the control signal, which is an input to the controlled flow network $v_\phi^C$. In the case of a deterministic simulator, this makes the control signal a function of $\boldsymbol{\theta}_t$. The controlled flow network is trained with the same loss as vanilla flow matching \citep{DBLP:conf/iclr/LipmanCBNL23}. Therefore all theoretical properties remain preserved.

\section{Simulation-based Inference}

\paraspace{}
\bh{This section is organized as follows. First, in section \ref{sec:sbi_tasks}, we introduce a set of SBI benchmark tasks and provide a comparison of 
popular neural posterior estimation (NPE) methods against 
a baseline of flow matching without simulator feedback.
This comparison uses a similar training setup for all models and tasks. 
Then, in section \ref{sec:training_variants}, we focus on an optimal task-specific network with training hyperparameters based on an extensive grid search. 
We evaluate different variants of flow matching that are related to simulator feedback on the SBI tasks to push the performance as far as possible.
In section \ref{sec:simulator_feedback}, we pick the most challenging SBI task and improve it further by introducing simulator feedback via gradient-based and learned control signals. We carefully analyze the cost-accuracy trade-off for using simulators and show that improvements from simulator feedback cannot be replicated by increasing the training dataset size alone.}

\paraspace{}
\subsection{Tasks and baselines} \label{sec:sbi_tasks}

\begin{wraptable}{r}{70mm}
\centering\vspace{-0.41cm}
    \captionof{table}{C2ST comparison with identical training setups %(learning rate, batch size, etc.) 
    and comparable number of network weights (ca. $300$K).\label{tab:npe}}
    \begin{tabular}{p{1.7cm}cccc}
      \bf Method    & \bf LV & \bf SLCP & \bf SIR & \bf TM \\ \hline \\
        CNF    & 0.99 & \underline{0.80} & 0.99 & 0.60 \\
        NSF   & 0.99  & - & \bf 0.75 & \bf 0.54 \\
        FFJORD  & \underline{0.95} & 0.82  & \underline{0.78} & 0.59 \\
        \textit{Flow-Mat.}    & \bf 0.93 & \bf 0.79 &  \underline{0.79} &\underline{0.58} \\ 
        %\textit{Flow-Mat.} & 0.53 & 0.61 & 0.84 & 0.90 \\
      \end{tabular}\vspace{-0.31cm}
\end{wraptable}

We consider the SBI tasks Lotka Volterra \textbf{LV}, a coupled ODE for the population dynamics of interacting species, \textbf{SIR}, an epidemiological model for the spread of diseases, \textbf{SLCP} and Two Moons (\textbf{TM}), two synthetic tasks having complicated multimodal posteriors. All tasks are part of the benchmark collection from \cite{DBLP:conf/aistats/LueckmannBGGM21}. 
For each problem, the posterior distribution for a set of $10$ observations is known, which allows for directly comparing it with the posterior predicted by the trained model. This is measured using the C2ST score \citep{DBLP:conf/iclr/Lopez-PazO17}, which trains a classifier to discriminate between samples from the true posterior and samples generated from the learned model. If the classifier cannot discriminate between two sets of samples, its test accuracy will be $0.5$, whereas it increases when they become more dissimilar. 

We include the following baseline methods for NPE: 
Continuous normalizing flows \citep[CNF]{DBLP:conf/nips/ChenRBD18},
Neural Splince Flows \citep[NSF]{DBLP:conf/nips/DurkanB0P19},
and 
FFJORD \citep{DBLP:conf/iclr/GrathwohlCBSD19}. 
Since we propose to include feedback from simulators, here we focus on the largest benchmark budget of $10^5$ simulator calls for generating the training dataset. Table~\ref{tab:npe} highlights that flow matching yields a highly competitive performance in this setting. 
For details on the training setup, see appendix \ref{sec:sbi}. 

Flow matching has also been evaluated for the SBI benchmark tasks by \cite{DBLP:conf/nips/WildbergerDBGMS23}, who performed an extensive hyperparamter search for each task to find optimal hyperparameters. 
In the following, we focus on flow matching, and hence use the corresponding sets of optimal hyperparameters for each task. 

\begin{figure}[t]
\centering
%\vspace*{-.5cm}
\includegraphics[width=.6\textwidth]{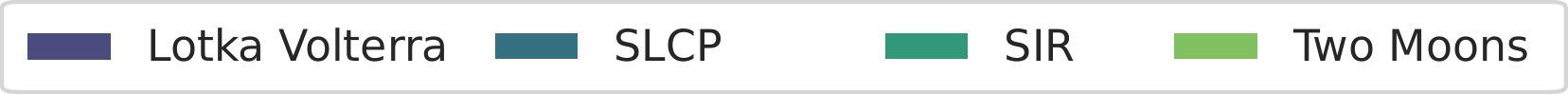}
 \\
\includegraphics[width=.36\textwidth]{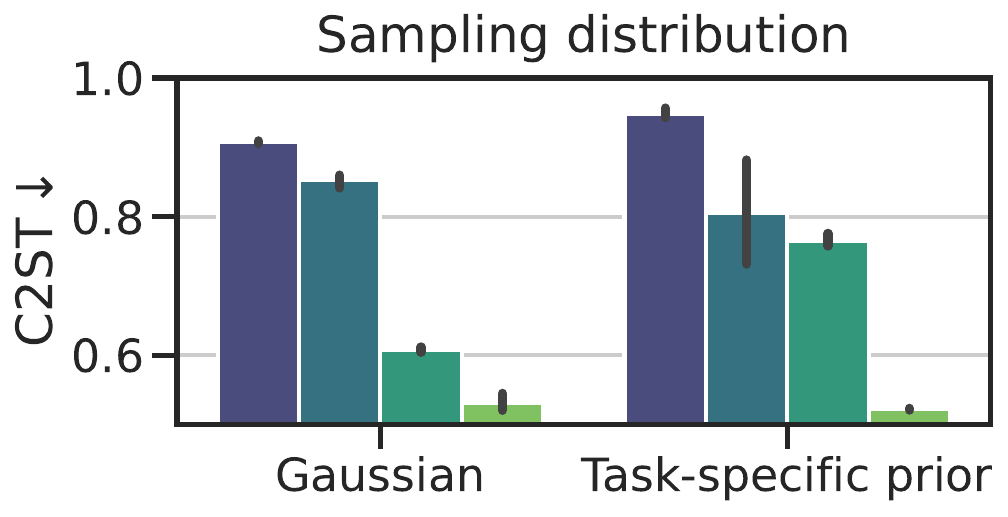}
\includegraphics[width=.30\textwidth]{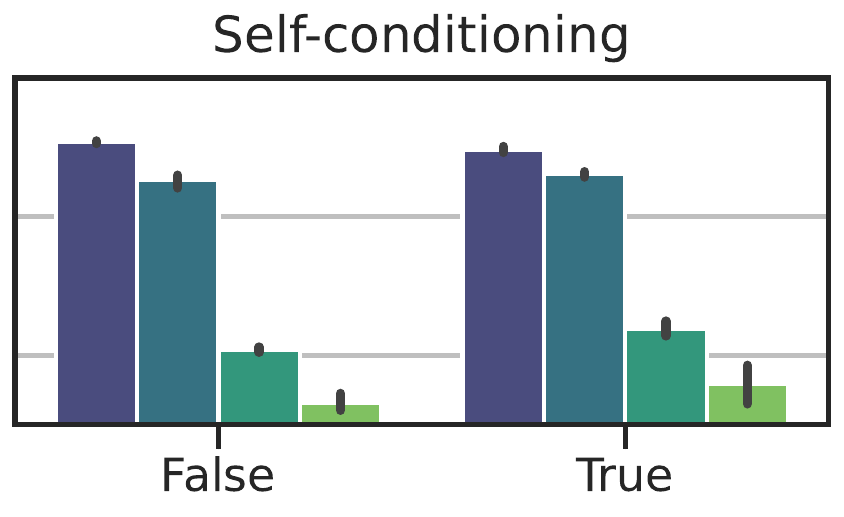}
\includegraphics[width=.30\textwidth]{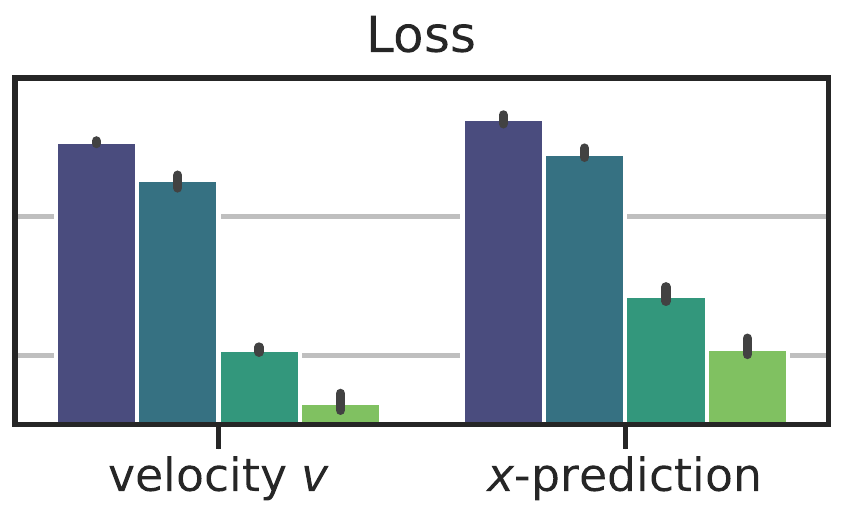}
\caption{
Evaluation of SBI tasks using different variants of flow matching training. Lower C2ST scores are better.}
\label{fig:sbi_benchmarks}
\end{figure}
\subsection{Training variants} \label{sec:training_variants}
There are several variants of training diffusion models that can be related to simulation feedback and which we consider promising in the context of SBI. Before we go on to evaluate the simulator feedback in section \ref{sec:simulator_feedback}, we test if we can improve the performance using any of them. In particular, we assess the following modifications: 
\begin{itemize}[leftmargin=10pt]
    \item \textbf{Self-conditioning}: conditioning a model on something that depends on its own output can be seen as a form of self-conditioning. We evaluate an adapted version of self-conditioning \citep{DBLP:conf/iclr/ChenZH23}. Instead of providing $\boldsymbol{\theta}_t$ to the flow network, the input is comprised of the concatenated vector \smash{$[\boldsymbol{\theta}_t; \mathrm{Dropout}(\hat{\boldsymbol{\theta}}_1)]$}, where \smash{$\hat{\boldsymbol{\theta}}_1$} is the 1-step prediction \eqref{eq:1_step_prediction}. For computing \smash{$\hat{\boldsymbol{\theta}}_1$}, we require one network evaluation with the input $[\boldsymbol{\theta}_t; 0]$ and stop the gradient backpropagation at \smash{$\hat{\boldsymbol{\theta}}_1$}. This method is similar to our simulator feedback, as it introduces a feedback loop that conditions the model on its own output, but without any simulator. 
    \item \textbf{Independent couplings}: It is also possible to couple two non-Gaussian distributions by defining the coupling as $q(\boldsymbol{z}) = p_0(\boldsymbol{\theta}_0) p_1(\boldsymbol{\theta}_1)$ and setting the conditional probabilities to the linear paths defined by $p_t(\boldsymbol{\theta}|(\boldsymbol{\theta}_0,\boldsymbol{\theta}_1)) = \mathcal{N}(\boldsymbol{\theta} | t\boldsymbol{\theta}_1 + (1-t)\boldsymbol{\theta}_0, \sigma I)$ and $u_t(\boldsymbol{\theta}|(\boldsymbol{\theta}_0,\boldsymbol{\theta}_1)) = \boldsymbol{\theta}_1 - \boldsymbol{\theta}_0$ with bandwidth $\sigma > 0$ \citep{DBLP:conf/iclr/LiuG023, DBLP:journals/corr/tong2023}. We can choose $p_0$ as the prior distribution $p(\boldsymbol{\theta})$ which we know in the SBI setting. Obtaining information in the form of an observation changes our knowledge about $\theta$ from the prior distribution to the posterior, therefore resembling a transformation similar to the noise to data transformation in diffusion models. This also suggests that the prior distribution can be closer to the posterior than a noise distribution.
    \item \textbf{$x$-prediction}: The reliability of the control signal depends directly on the 1-step estimate \smash{$\hat{\boldsymbol{\theta}}$}. Instead of regressing the flow $u_t(\theta)$, we can directly predict the denoised estimate \smash{$\hat{\boldsymbol{\theta}}$} and obtain the velocity by rearranging \eqref{eq:1_step_prediction}, giving \smash{$v_\phi(t, \boldsymbol{\theta}_t, \boldsymbol{x}_o) = \hat{\boldsymbol{\theta}}_1/(1-t)$}. We additionally weight the $x$-prediction loss with a time-dependent weighting $w_t := 1/(1-t)$ to account for the scaling in \eqref{eq:1_step_prediction}. The $x$-prediction potentially produces better estimates for \smash{$\hat{\boldsymbol{\theta}}$}, thus allowing for obtaining more reliable feedback from control signals when $t < 0.8$.   
\end{itemize}
\paraspace{}\paragraph{Evaluation} Figure \ref{fig:sbi_benchmarks} shows an evaluation of the different variants against vanilla flow matching (Gaussian sampling distribution, no self-conditioning and velocity prediction). Using task-specific priors produces outliers with better C2ST scores for \text{SLCP} but is consistently worse for LV and SIR. We conclude that normal Gaussian distributions are more suited as sampling distributions for most low-dimensional problems. Introducing self-conditioning does not show any improvements, so feedback loops without a simulator alone are not sufficient for better performance in this situation. Finally, the $x$-prediction loss consistently performs worse than the velocity prediction. Therefore, a potential improvement in the 1-step estimate is outweighed by a corresponding deterioration of the posterior correctness as indicated by the C2ST score.
\newpage
\subsection{Simulator feedback: Gradient-based and Learned}
\label{sec:simulator_feedback}
\begin{wrapfigure}{r}{.4\textwidth}
\vspace*{-.5cm}
\centering
\includegraphics[width=.4\textwidth]{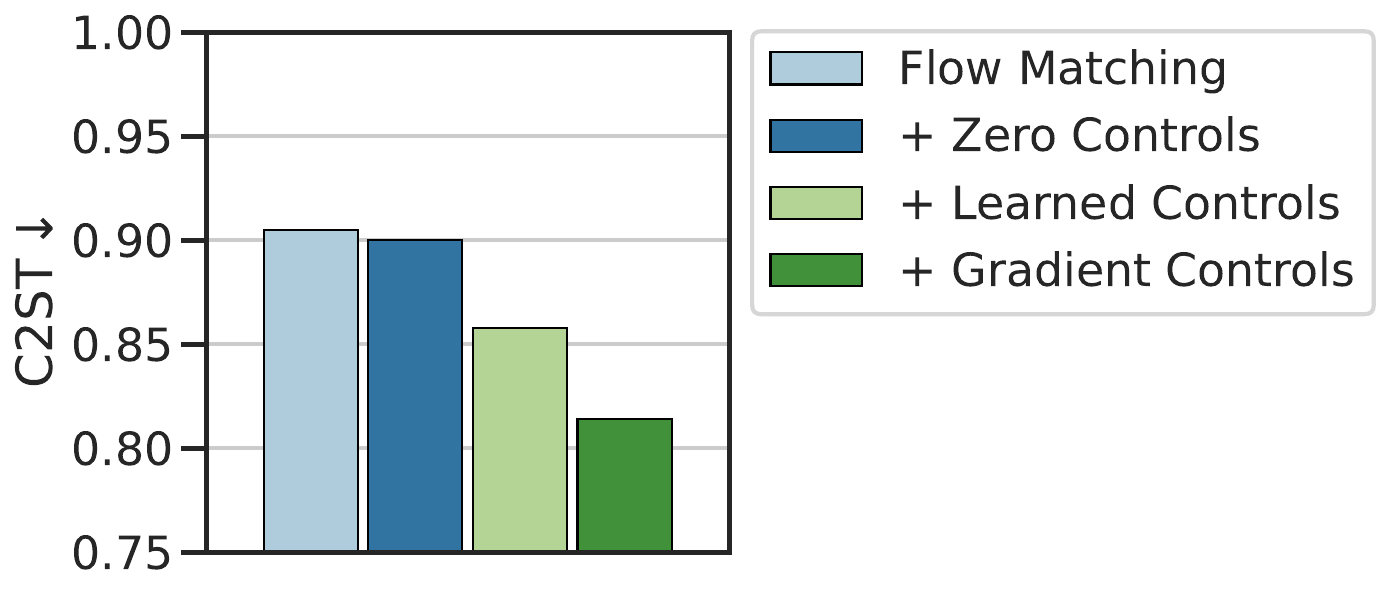}
\caption{Evaluation of simulator feedback for \textbf{LV}. }
\label{fig:simulator_feedback_eval}
\centering
\includegraphics[width=.4\textwidth]{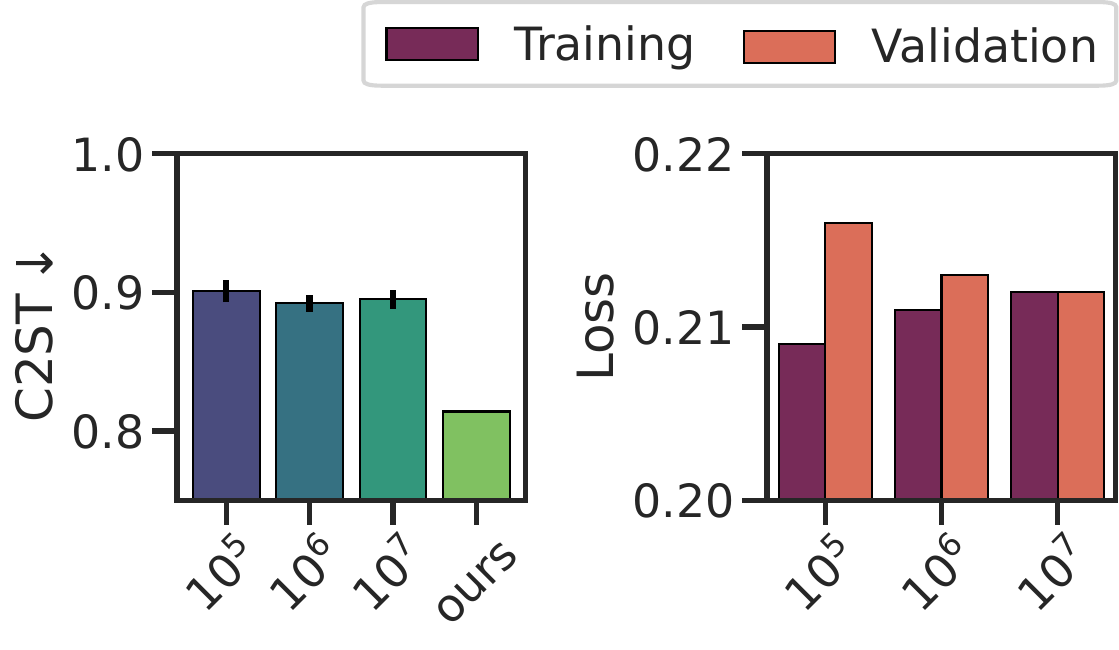}
\caption{Different simulator call budgets (training set sizes $10^5$, $10^6$, $10^7$) compared with finetuning using simulator feedback (ours, ca. $9 \times 10^6$ simulator calls in total). }
\label{fig:computational_efficiency}
\end{wrapfigure}
\paraspace{} In this section, we focus on the Lotka-Volterra (LV) task for a more detailed analysis. It has the highest difficulty as seen by the C2ST score, and we use it to test the different types of feedback. We reimplement the LV simulator in JAX \citep{jax2018github} to support differentiability and evaluate the gradient-based control signal as well as the learning-based control signal, using a small multilayer perceptron (MLP). In addition, to make sure that observed improvements are not due to the increased number of network parameters and finetuning with the control network, we also evaluate a variant where we finetune with the control network but set all simulator-dependent inputs to the control network to $0$ (Zero Controls). 
We show an evaluation with C2ST in \figref{fig:simulator_feedback_eval}. For both the learning and gradient-based control signals we see clear improvements with the gradient-based signal clearly ahead. The zero control signal improves only slightly, showing that the improvement can be directly attributed to the simulator.
While control signals are most useful for more high-dimensional problems with less sparse and noisy observations, this experiment demonstrates that they can also be used in low-dimensional settings. Moreover, while differentiable simulators can provide better control signals, feedback from non-differentiable simulators likewise shows clear improvements. 
\subsection{Computational efficiency}
\paraspace{} A critical issue in SBI is that calls to the simulator are potentially expensive. This imposes the question of whether compute time is better spent on extending the training dataset or training with feedback from the simulator. We empirically verify that the latter is more efficient for the LV task in this setup by comparing our method to models with an increased training dataset from a larger simulator budget. Specifically, we train with dataset sizes of $10^6$ and $10^7$. Training the gradient-based control signal took ca. $9 \times 10^6$ simulator calls. See \figref{fig:computational_efficiency} for the evaluation. There is no improvement in the C2ST for models trained without simulator feedback beyond $10^5$ data points, and the final train/validation loss for the $10^7$ model indicates that there is no more overfitting.
Nonetheless, the model trained with controls clearly outperforms the model trained with more data, indicating that the directed feedback of the simulator cannot be replaced by increased amounts of training data.   
\section{Strong Gravitational Lensing} \label{sec:strong_gravitational_lensing}
\paraspace{} We present our results for modeling strong gravitational lens systems, a challenging and highly relevant non-linear problem in astronomy.  
Strong gravitational lensing is a physical phenomenon whereby the light rays by a distant object, such as a galaxy, are deflected by an intervening massive object, such as another galaxy or a galaxy cluster. As a result, one observes multiple distorted images of the background source. We aim to recover both the lens and source light distribution as well as the lens mass density distribution with realistic simulated observations for which we know the ground truths. We evaluate flow matching as an NPE method with gradient-based control signals from a differentiable simulator with two MCMC methods.
\paraspace{} \paragraph{Lens modeling} The \textit{lens equation} relates coordinates on the source plane $\beta$ and the observed image plane $\Theta$ via the deflection angle $\alpha$ induced by the mass profile or gravitational potential of the lens galaxy. We use a Singular Isothermal Ellipsoid (SIE) to describe this lens mass and Sérsic profiles for both the source light and light emitted from the lens galaxy (full details are provided in appendix \ref{app:strong_lensing}). There are $9$ parameters for the lens mass and $7$ parameters for each Sérsic profile, giving $23$ parameters in total. The likelihood is measured by the $\chi^2$-statistic, which is the modeled image plane $\Theta$ minus the observation $\boldsymbol{x}_o$ divided by the noise. To solve the lensing equation, we make use of the publicly available raytracing code by \citep{herculens2022}. We want to stress that even small perturbations of the model parameters can cause the $\chi_2$ to increase significantly; see  \figref{fig:noise_reconstruction} in the appendix.
\myskip
\begin{figure}[t]
    \centering
    \includegraphics[width=\textwidth]{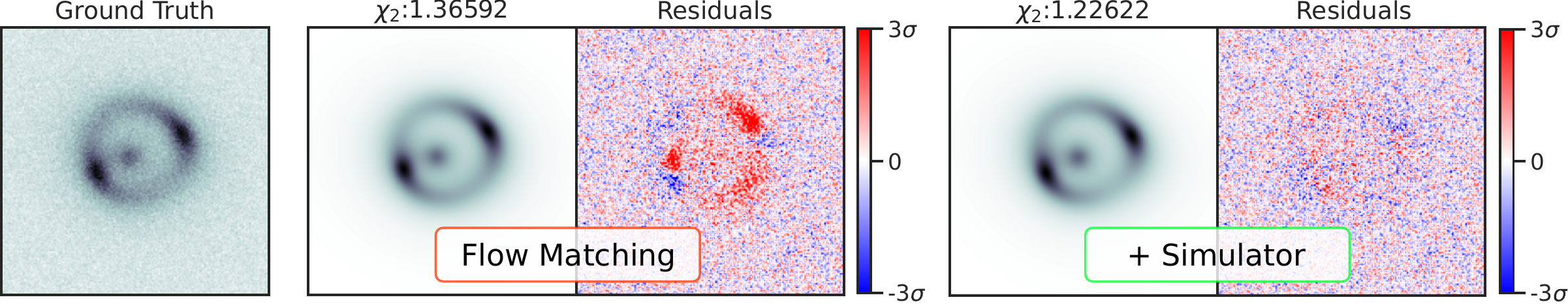}
    \caption{Reconstruction of observation. Flow matching is purely learning-based and shows noticeable residuals in the reconstruction. Including simulator feedback removes remaining residuals.}
    \label{fig:lensing}
\end{figure}
\paraspace{}\paragraph{Datasets and pretraining}
Several instrument-specific measurement effects are included when simulating the observations. We include background and Poisson noise and smoothing by a point-spread function (PSF). The pixel size corresponds to $0.04$ arc seconds. These directly affect the posterior, as more noise and a stronger PSF will widen the posterior distribution. We generate 250 000 data samples for training and 25 000 for validation. The flow network $v_\phi$ consists of a convolutional feature extraction neural network represented by a shallow CNN whose output is fed into a dense feed-forward neural network with residual blocks. Full details are in appendix \ref{app:strong_lensing}.
\paraspace{}\paragraph{Finetuning with control signals}
The control network $v_\phi^C$ is represented by another dense feed-forward network, which accounts for $11\%$ of all parameters in the combined model. The control signals are obtained from simulating an observation based on the predicted estimate \smash{$\hat{\boldsymbol{\theta}}_1$} via ray-tracing \citep{herculens2022} based on the parametric models, calculating the $\chi_2$-statistic and computing gradients with respect to the estimate \smash{$\hat{\boldsymbol{\theta}}_1$}. The $\chi_2$-statistic itself is also part of the control signal.
\paraspace{} \paragraph{Reference posteriors} As reference posteriors, we include Hamiltonian Monte Carlo (HMC) with No-U-Turn sampler \citep[NUTS]{hoffman2014no} and Affine-Invariant Ensemble Sampling \citep[AIES]{goodman2010ensemble}, which are both two popular MCMC-methods in astronomy. We adopt implementations of both methods using numpyro \citep{phan2019composable, bingham2019pyro}. 
Details on running both methods can be found in appendix \ref{app:strong_lensing}. Additionally, we compare to diffusion posterior sampling \nt{as a learned baseline \citep[DPS]{DBLP:conf/nips/ChungKY23}, see appendix \ref{sec:dps} for details. 
We use Euler integration for both flow matching variants.} 
\subsection{Evaluation and Discussion}
\begin{wraptable}{r}{70mm}
\centering\vspace{-0.41cm}
\caption{Evaluation with respect to average $\chi_2$ and inference time for the posterior distribution.}
\label{tab:lensing_evaluation}
\begin{tabular}{lll}
            \bf Method & \bf Avg. $\chi_2 \downarrow$ & \bf Modeling Time $\downarrow$ \\
            \hline \\  
            NUTS & 1.83 & $\sim$ 56x (564s) \\
            AIES & 1.74 & $\sim$ 67x (672s) 
            \\ & \\
            DPS & 9.98 & $\sim$ 42x (427s)
            \\ & \\
            \emph{Flow-Mat.}  & 1.83 & 1x (\textbf{10s}) \\
            + Simulator & \bf 1.48 & $\sim$ 2x (19s) \\
          \end{tabular}
\end{wraptable}
\paraspace{}\paragraph{$\chi_2$-statistic} We show an evaluation of all methods in table \ref{tab:lensing_evaluation}. The average $\chi_2$ is computed over $1 000$ randomly chosen validation systems, where for each, we draw $1 000$ samples from the posterior. 
If we compute the $\chi_2$ for the ground truth parameters, we obtain a value of $1.17$ due to the noise in the observation. Since we cannot overfit to noise with the parametric models, this represents a lower bound for $\chi_2$ in this experiment.   
Including the physics-based control improves the $\chi_2$ from $1.83$ to $1.48$, representing an improvement of $53\%$ relative to the best modeling. The improved $\chi_2$ is even better than the best baseline method, AIES.
\paraspace{} \paragraph{Modeling time} We define the modeling time as the average compute time required to produce $1 000$ credible posterior samples. Both HMC and AIES require significant warmup times before producing the first samples from the posterior, which we include in the table. However, after warmup, it is relatively cheap to obtain new samples. On the other hand, flow matching does not require any warmup time and the modeling time increases linearly with the number of posterior samples. All methods were implemented in JAX \citep{jax2018github} and used the same hardware.  
The measurements in table \ref{tab:lensing_evaluation} \nt{show that DPS is faster than the classic baselines, but yields a very sub-optimal performance in terms of its distribution.
The performance numbers} also highlight that our method yields an accuracy that surpasses AIES, while being more than 30x faster. 
This evaluation demonstrates that \nt{flow matching-based} 
methods are  highly competitive  even in small to moderate-sized problems 
where established MCMC methods in terms of accuracy exist, clearly beating them in terms of inference time. 
Flow matching with our proposed control signals is especially interesting because it is not affected as much by the curse of dimensionality as traditional inference methods and allows for having non-trivial learnable high-dimensional priors. 
However, before these methods are widely trusted, 
they need to demonstrate their competitiveness with classical methods. 
Our results show that this is indeed the case, which opens up exciting avenues for applying and developing approaches targeting similar and adjacent inverse problems in science. 

\begin{wrapfigure}{r}{.36\textwidth}
\centering\vspace{-11pt}
\includegraphics[width=.36\textwidth]{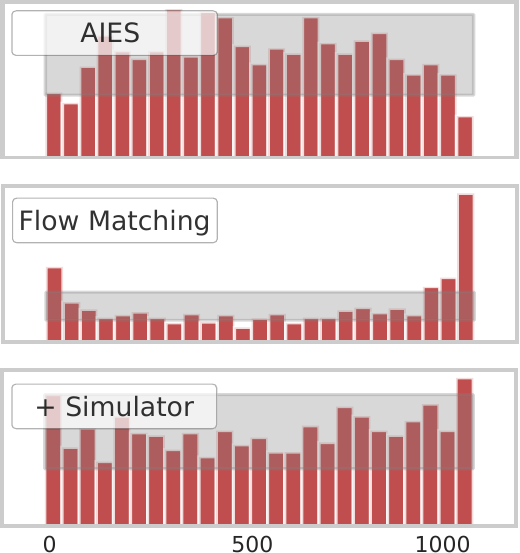}
\caption{SBC for $x_\mathrm{center}$ of the source galaxy. }\vspace{-5pt}
\label{fig:simulation-based-calibration}
\end{wrapfigure}
\paragraph{Simulation-based calibration}
\paraspace{} Acquiring truthful posterior distributions for Bayesian inference problems is difficult, which makes it hard to robustly evaluate whether the predicted posterior distribution is correct. We use simulation-based calibration \citep[SBC]{talts2018validating} as an additional evaluation tool. The data-averaged posterior obtained from averaging the posterior distribution over many problem instances has to be equal to the prior. This can be tested by considering a one-dimensional function $f: \boldsymbol{\theta} \mapsto \mathbb{R}$ and $L$ samples $\boldsymbol{\theta}^1,..., \boldsymbol{\theta}^L$ drawn from an inference method. If $\boldsymbol{\theta}^*$ are the ground truth parameters, then the rank statistic $\sum_{l=1}^L \mathbf{1}_{f(\boldsymbol{\theta}^L) < f(\boldsymbol{\theta}^*)}$ has to be uniformly distributed over the integers $[0,L]$. If the distribution of the rank statistic is plotted as a histogram, systematic problems in the inference method can be identified visually, see \figref{fig:simulation-based-calibration}. We set $L=1000$ and plot the histograms for all $n=1000$ test problems and visualize the parameter $x_\mathrm{center}$, which defines the position of the source in $x$-direction. The posteriors without simulator feedback are biased, as can be seen in the deviation from uniformity in the plots. Including simulator feedback improves the distribution of the rank statistic. For an extended analysis, see appendix \ref{sec:sbc}.
\subsection{Limitations}
\paraspace{} While introducing additional control signals increases the quality of produced samples, it comes at the cost of slower inference and training times depending on the speed of the simulator.
In general, using non-differentiable control signals is possible but removes the possibility of computing likelihoods via the instantaneous change of variables formula \citep{DBLP:conf/nips/ChenRBD18}. Compared to MCMC approaches, inference with flow-based models requires a substantial upfront cost for training that needs to be amortized across many problems. Additionally, priors are encoded in the learned flow networks, so changing them would require retraining models with adjusted data sets.
\paraspace{}
\section{Conclusion}
\paraspace{} We presented a method for improving flow-based models with simulator feedback using control signals. 
This allows us to refine an existing flow with only a few additional weights and little training time. We thereby efficiently bridge the gap between purely learning-based methods for simulation-based inference and optimization with hand-crafted cost functions within the framework of flow matching. This improvement is critical for scientific applications where high accuracy and trustworthiness in the methods are required. Purely learning-based methods face significant difficulties in producing very accurate samples, as there is usually no feedback during inference of how good samples are. 
In this paper, we demonstrated that we do not need large network sizes or tremendous amounts of data to train accurate models that are competitive with established MCMC methods if we include suitable control signals from simulators. We believe this work makes an important step towards making posterior inference in science more accurate, understandable, and reliable. 

\bibliography{iclr2025_conference}

\begin{thebibliography}{54}
\providecommand{\natexlab}[1]{#1}
\providecommand{\url}[1]{\texttt{#1}}
\expandafter\ifx\csname urlstyle\endcsname\relax
  \providecommand{\doi}[1]{doi: #1}\else
  \providecommand{\doi}{doi: \begingroup \urlstyle{rm}\Url}\fi

\bibitem[Albergo et~al.(2023{\natexlab{a}})Albergo, Boffi, and Vanden{-}Eijnden]{DBLP:journals/corr/albergo2023}
Michael~S. Albergo, Nicholas~M. Boffi, and Eric Vanden{-}Eijnden.
\newblock Stochastic interpolants: {A} unifying framework for flows and diffusions.
\newblock \emph{CoRR}, abs/2303.08797, 2023{\natexlab{a}}.
\newblock \doi{10.48550/ARXIV.2303.08797}.
\newblock URL \url{https://doi.org/10.48550/arXiv.2303.08797}.

\bibitem[Albergo et~al.(2023{\natexlab{b}})Albergo, Goldstein, Boffi, Ranganath, and Vanden{-}Eijnden]{DBLP:journals/corr/albergo2023coupling}
Michael~S. Albergo, Mark Goldstein, Nicholas~M. Boffi, Rajesh Ranganath, and Eric Vanden{-}Eijnden.
\newblock Stochastic interpolants with data-dependent couplings.
\newblock \emph{CoRR}, abs/2310.03725, 2023{\natexlab{b}}.
\newblock \doi{10.48550/ARXIV.2310.03725}.
\newblock URL \url{https://doi.org/10.48550/arXiv.2310.03725}.

\bibitem[Baydin et~al.(2019)Baydin, Shao, Bhimji, Heinrich, Naderiparizi, Munk, Liu, Gram{-}Hansen, Louppe, Meadows, Torr, Lee, Cranmer, Prabhat, and Wood]{DBLP:conf/nips/BaydinSBHNM0GLM19}
Atilim~Gunes Baydin, Lei Shao, Wahid Bhimji, Lukas Heinrich, Saeid Naderiparizi, Andreas Munk, Jialin Liu, Bradley Gram{-}Hansen, Gilles Louppe, Lawrence Meadows, Philip H.~S. Torr, Victor~W. Lee, Kyle Cranmer, Prabhat, and Frank Wood.
\newblock Efficient probabilistic inference in the quest for physics beyond the standard model.
\newblock In \emph{Advances in Neural Information Processing Systems 32}, pp.\  5460--5473, 2019.
\newblock URL \url{https://proceedings.neurips.cc/paper/2019/hash/6d19c113404cee55b4036fce1a37c058-Abstract.html}.

\bibitem[Bingham et~al.(2019)Bingham, Chen, Jankowiak, Obermeyer, Pradhan, Karaletsos, Singh, Szerlip, Horsfall, and Goodman]{bingham2019pyro}
Eli Bingham, Jonathan~P. Chen, Martin Jankowiak, Fritz Obermeyer, Neeraj Pradhan, Theofanis Karaletsos, Rohit Singh, Paul~A. Szerlip, Paul Horsfall, and Noah~D. Goodman.
\newblock Pyro: Deep universal probabilistic programming.
\newblock \emph{J. Mach. Learn. Res.}, 20:\penalty0 28:1--28:6, 2019.
\newblock URL \url{http://jmlr.org/papers/v20/18-403.html}.

\bibitem[Bradbury et~al.(2018)Bradbury, Frostig, Hawkins, Johnson, Leary, Maclaurin, Necula, Paszke, Vander{P}las, Wanderman-{M}ilne, and Zhang]{jax2018github}
James Bradbury, Roy Frostig, Peter Hawkins, Matthew~James Johnson, Chris Leary, Dougal Maclaurin, George Necula, Adam Paszke, Jake Vander{P}las, Skye Wanderman-{M}ilne, and Qiao Zhang.
\newblock {JAX}: composable transformations of {P}ython+{N}um{P}y programs, 2018.
\newblock URL \url{http://github.com/google/jax}.

\bibitem[Cardoso et~al.(2023)Cardoso, Idrissi, Corff, and Moulines]{DBLP:journals/corr/cardoso2023}
Gabriel Cardoso, Yazid Janati~El Idrissi, Sylvain~Le Corff, and Eric Moulines.
\newblock Monte carlo guided diffusion for bayesian linear inverse problems.
\newblock \emph{CoRR}, abs/2308.07983, 2023.
\newblock \doi{10.48550/ARXIV.2308.07983}.
\newblock URL \url{https://doi.org/10.48550/arXiv.2308.07983}.

\bibitem[Chen et~al.(2018)Chen, Rubanova, Bettencourt, and Duvenaud]{DBLP:conf/nips/ChenRBD18}
Tian~Qi Chen, Yulia Rubanova, Jesse Bettencourt, and David Duvenaud.
\newblock Neural ordinary differential equations.
\newblock In \emph{Advances in Neural Information Processing Systems 31}, pp.\  6572--6583, 2018.
\newblock URL \url{https://proceedings.neurips.cc/paper/2018/hash/69386f6bb1dfed68692a24c8686939b9-Abstract.html}.

\bibitem[Chen et~al.(2023)Chen, Zhang, and Hinton]{DBLP:conf/iclr/ChenZH23}
Ting Chen, Ruixiang Zhang, and Geoffrey~E. Hinton.
\newblock Analog bits: Generating discrete data using diffusion models with self-conditioning.
\newblock In \emph{The Eleventh International Conference on Learning Representations, {ICLR} 2023, Kigali, Rwanda, May 1-5, 2023}. OpenReview.net, 2023.
\newblock URL \url{https://openreview.net/pdf?id=3itjR9QxFw}.

\bibitem[Chung et~al.(2022)Chung, Sim, and Ye]{DBLP:conf/cvpr/ChungSY22}
Hyungjin Chung, Byeongsu Sim, and Jong~Chul Ye.
\newblock Come-closer-diffuse-faster: Accelerating conditional diffusion models for inverse problems through stochastic contraction.
\newblock In \emph{{IEEE/CVF} Conference on Computer Vision and Pattern Recognition, {CVPR} 2022, New Orleans, LA, USA, June 18-24, 2022}, pp.\  12403--12412. {IEEE}, 2022.
\newblock \doi{10.1109/CVPR52688.2022.01209}.
\newblock URL \url{https://doi.org/10.1109/CVPR52688.2022.01209}.

\bibitem[Chung et~al.(2023{\natexlab{a}})Chung, Kim, McCann, Klasky, and Ye]{DBLP:conf/iclr/ChungKMKY23}
Hyungjin Chung, Jeongsol Kim, Michael~Thompson McCann, Marc~Louis Klasky, and Jong~Chul Ye.
\newblock Diffusion posterior sampling for general noisy inverse problems.
\newblock In \emph{The Eleventh International Conference on Learning Representations, {ICLR} 2023, Kigali, Rwanda, May 1-5, 2023}. OpenReview.net, 2023{\natexlab{a}}.
\newblock URL \url{https://openreview.net/pdf?id=OnD9zGAGT0k}.

\bibitem[Chung et~al.(2023{\natexlab{b}})Chung, Kim, and Ye]{DBLP:conf/nips/ChungKY23}
Hyungjin Chung, Jeongsol Kim, and Jong~Chul Ye.
\newblock Direct diffusion bridge using data consistency for inverse problems.
\newblock In \emph{Advances in Neural Information Processing Systems 36}, 2023{\natexlab{b}}.
\newblock URL \url{http://papers.nips.cc/paper\_files/paper/2023/hash/165b0e600b1721bd59526131eb061092-Abstract-Conference.html}.

\bibitem[Cranmer et~al.(2020)Cranmer, Brehmer, and Louppe]{cranmer2020frontier}
Kyle Cranmer, Johann Brehmer, and Gilles Louppe.
\newblock The frontier of simulation-based inference.
\newblock \emph{Proceedings of the National Academy of Sciences}, 117\penalty0 (48):\penalty0 30055--30062, 2020.

\bibitem[Cunha \& Herdeiro(2018)Cunha and Herdeiro]{cunha2018shadows}
Pedro~VP Cunha and Carlos~AR Herdeiro.
\newblock Shadows and strong gravitational lensing: a brief review.
\newblock \emph{General Relativity and Gravitation}, 50:\penalty0 1--27, 2018.

\bibitem[Dax et~al.(2021)Dax, Green, Gair, Macke, Buonanno, and Sch{\"o}lkopf]{dax2021real}
Maximilian Dax, Stephen~R Green, Jonathan Gair, Jakob~H Macke, Alessandra Buonanno, and Bernhard Sch{\"o}lkopf.
\newblock Real-time gravitational wave science with neural posterior estimation.
\newblock \emph{Physical review letters}, 127\penalty0 (24):\penalty0 241103, 2021.

\bibitem[Dinh et~al.(2017)Dinh, Sohl{-}Dickstein, and Bengio]{DBLP:conf/iclr/DinhSB17}
Laurent Dinh, Jascha Sohl{-}Dickstein, and Samy Bengio.
\newblock Density estimation using real {NVP}.
\newblock In \emph{5th International Conference on Learning Representations, {ICLR} 2017, Toulon, France, April 24-26, 2017, Conference Track Proceedings}. OpenReview.net, 2017.
\newblock URL \url{https://openreview.net/forum?id=HkpbnH9lx}.

\bibitem[Durkan et~al.(2019)Durkan, Bekasov, Murray, and Papamakarios]{DBLP:conf/nips/DurkanB0P19}
Conor Durkan, Artur Bekasov, Iain Murray, and George Papamakarios.
\newblock Neural spline flows.
\newblock In \emph{Advances in Neural Information Processing Systems 32}, pp.\  7509--7520, 2019.
\newblock URL \url{https://proceedings.neurips.cc/paper/2019/hash/7ac71d433f282034e088473244df8c02-Abstract.html}.

\bibitem[{Galan} et~al.(2022){Galan}, {Vernardos}, {Peel}, {Courbin}, and {Starck}]{herculens2022}
A.~{Galan}, G.~{Vernardos}, A.~{Peel}, F.~{Courbin}, and J.~L. {Starck}.
\newblock {Using wavelets to capture deviations from smoothness in galaxy-scale strong lenses}.
\newblock \emph{Astronomy and Astrophysics}, 668:\penalty0 A155, December 2022.
\newblock \doi{10.1051/0004-6361/202244464}.

\bibitem[Gneiting \& Raftery(2005)Gneiting and Raftery]{gneiting2005weather}
Tilmann Gneiting and Adrian~E Raftery.
\newblock Weather forecasting with ensemble methods.
\newblock \emph{Science}, 310\penalty0 (5746):\penalty0 248--249, 2005.

\bibitem[Goodman \& Weare(2010)Goodman and Weare]{goodman2010ensemble}
Jonathan Goodman and Jonathan Weare.
\newblock Ensemble samplers with affine invariance.
\newblock \emph{Communications in applied mathematics and computational science}, 5\penalty0 (1):\penalty0 65--80, 2010.

\bibitem[Grathwohl et~al.(2019)Grathwohl, Chen, Bettencourt, Sutskever, and Duvenaud]{DBLP:conf/iclr/GrathwohlCBSD19}
Will Grathwohl, Ricky T.~Q. Chen, Jesse Bettencourt, Ilya Sutskever, and David Duvenaud.
\newblock {FFJORD:} free-form continuous dynamics for scalable reversible generative models.
\newblock In \emph{7th International Conference on Learning Representations, {ICLR} 2019, New Orleans, LA, USA, May 6-9, 2019}. OpenReview.net, 2019.
\newblock URL \url{https://openreview.net/forum?id=rJxgknCcK7}.

\bibitem[Gretton et~al.(2012)Gretton, Borgwardt, Rasch, Sch{\"o}lkopf, and Smola]{gretton2012kernel}
Arthur Gretton, Karsten~M Borgwardt, Malte~J Rasch, Bernhard Sch{\"o}lkopf, and Alexander Smola.
\newblock A kernel two-sample test.
\newblock \emph{The Journal of Machine Learning Research}, 13\penalty0 (1):\penalty0 723--773, 2012.

\bibitem[Hezaveh et~al.(2017)Hezaveh, Levasseur, and Marshall]{hezaveh2017fast}
Yashar~D Hezaveh, Laurence~Perreault Levasseur, and Philip~J Marshall.
\newblock Fast automated analysis of strong gravitational lenses with convolutional neural networks.
\newblock \emph{Nature}, 548\penalty0 (7669):\penalty0 555--557, 2017.

\bibitem[Ho \& Salimans(2022)Ho and Salimans]{DBLP:journals/corr/hoclassifierfree2022}
Jonathan Ho and Tim Salimans.
\newblock Classifier-free diffusion guidance.
\newblock \emph{CoRR}, abs/2207.12598, 2022.
\newblock \doi{10.48550/ARXIV.2207.12598}.
\newblock URL \url{https://doi.org/10.48550/arXiv.2207.12598}.

\bibitem[Ho et~al.(2020)Ho, Jain, and Abbeel]{DBLP:conf/nips/HoJA20}
Jonathan Ho, Ajay Jain, and Pieter Abbeel.
\newblock Denoising diffusion probabilistic models.
\newblock In \emph{Advances in Neural Information Processing Systems 33}, 2020.
\newblock URL \url{https://proceedings.neurips.cc/paper/2020/hash/4c5bcfec8584af0d967f1ab10179ca4b-Abstract.html}.

\bibitem[Hoffman et~al.(2014)Hoffman, Gelman, et~al.]{hoffman2014no}
Matthew~D Hoffman, Andrew Gelman, et~al.
\newblock The no-u-turn sampler: adaptively setting path lengths in hamiltonian monte carlo.
\newblock \emph{J. Mach. Learn. Res.}, 15\penalty0 (1):\penalty0 1593--1623, 2014.

\bibitem[Holzschuh et~al.(2023)Holzschuh, Vegetti, and Thuerey]{DBLP:conf/nips/HolzschuhVT23}
Benjamin~J. Holzschuh, Simona Vegetti, and Nils Thuerey.
\newblock Solving inverse physics problems with score matching.
\newblock In \emph{Advances in Neural Information Processing Systems 36}, 2023.
\newblock URL \url{http://papers.nips.cc/paper\_files/paper/2023/hash/c2f2230abc7ccf669f403be881d3ffb7-Abstract-Conference.html}.

\bibitem[Kawar et~al.(2021)Kawar, Vaksman, and Elad]{DBLP:conf/nips/KawarVE21}
Bahjat Kawar, Gregory Vaksman, and Michael Elad.
\newblock {SNIPS:} solving noisy inverse problems stochastically.
\newblock In \emph{Advances in Neural Information Processing Systems 34}, pp.\  21757--21769, 2021.
\newblock URL \url{https://proceedings.neurips.cc/paper/2021/hash/b5c01503041b70d41d80e3dbe31bbd8c-Abstract.html}.

\bibitem[Kawar et~al.(2022)Kawar, Elad, Ermon, and Song]{DBLP:conf/nips/KawarEES22}
Bahjat Kawar, Michael Elad, Stefano Ermon, and Jiaming Song.
\newblock Denoising diffusion restoration models.
\newblock In \emph{Advances in Neural Information Processing Systems 35}, 2022.
\newblock URL \url{http://papers.nips.cc/paper\_files/paper/2022/hash/95504595b6169131b6ed6cd72eb05616-Abstract-Conference.html}.

\bibitem[Kingma \& Ba(2015)Kingma and Ba]{DBLP:journals/corr/KingmaB14}
Diederik~P. Kingma and Jimmy Ba.
\newblock Adam: {A} method for stochastic optimization.
\newblock In Yoshua Bengio and Yann LeCun (eds.), \emph{3rd International Conference on Learning Representations, {ICLR} 2015, San Diego, CA, USA, May 7-9, 2015, Conference Track Proceedings}, 2015.
\newblock URL \url{http://arxiv.org/abs/1412.6980}.

\bibitem[Kingma \& Dhariwal(2018)Kingma and Dhariwal]{DBLP:conf/nips/KingmaD18}
Diederik~P. Kingma and Prafulla Dhariwal.
\newblock Glow: Generative flow with invertible 1x1 convolutions.
\newblock In \emph{Advances in Neural Information Processing Systems 31}, pp.\  10236--10245, 2018.
\newblock URL \url{https://proceedings.neurips.cc/paper/2018/hash/d139db6a236200b21cc7f752979132d0-Abstract.html}.

\bibitem[Laureijs et~al.(2011)Laureijs, Amiaux, Arduini, Augueres, Brinchmann, Cole, Cropper, Dabin, Duvet, Ealet, et~al.]{laureijs2011euclid}
Rene Laureijs, J~Amiaux, S~Arduini, J-L Augueres, J~Brinchmann, R~Cole, M~Cropper, C~Dabin, L~Duvet, A~Ealet, et~al.
\newblock Euclid definition study report.
\newblock \emph{arXiv preprint arXiv:1110.3193}, 2011.

\bibitem[Legin et~al.(2021)Legin, Hezaveh, Levasseur, and Wandelt]{legin2021simulation}
Ronan Legin, Yashar Hezaveh, Laurence~Perreault Levasseur, and Benjamin Wandelt.
\newblock Simulation-based inference of strong gravitational lensing parameters.
\newblock \emph{arXiv preprint arXiv:2112.05278}, 2021.

\bibitem[Legin et~al.(2023)Legin, Hezaveh, Perreault-Levasseur, and Wandelt]{legin2023framework}
Ronan Legin, Yashar Hezaveh, Laurence Perreault-Levasseur, and Benjamin Wandelt.
\newblock A framework for obtaining accurate posteriors of strong gravitational lensing parameters with flexible priors and implicit likelihoods using density estimation.
\newblock \emph{The Astrophysical Journal}, 943\penalty0 (1):\penalty0 4, 2023.

\bibitem[Levasseur et~al.(2017)Levasseur, Hezaveh, and Wechsler]{levasseur2017uncertainties}
Laurence~Perreault Levasseur, Yashar~D Hezaveh, and Risa~H Wechsler.
\newblock Uncertainties in parameters estimated with neural networks: Application to strong gravitational lensing.
\newblock \emph{The Astrophysical Journal Letters}, 850\penalty0 (1):\penalty0 L7, 2017.

\bibitem[Lipman et~al.(2023)Lipman, Chen, Ben{-}Hamu, Nickel, and Le]{DBLP:conf/iclr/LipmanCBNL23}
Yaron Lipman, Ricky T.~Q. Chen, Heli Ben{-}Hamu, Maximilian Nickel, and Matthew Le.
\newblock Flow matching for generative modeling.
\newblock In \emph{The Eleventh International Conference on Learning Representations, {ICLR} 2023, Kigali, Rwanda, May 1-5, 2023}. OpenReview.net, 2023.
\newblock URL \url{https://openreview.net/pdf?id=PqvMRDCJT9t}.

\bibitem[Liu et~al.(2023)Liu, Gong, and Liu]{DBLP:conf/iclr/LiuG023}
Xingchao Liu, Chengyue Gong, and Qiang Liu.
\newblock Flow straight and fast: Learning to generate and transfer data with rectified flow.
\newblock In \emph{The Eleventh International Conference on Learning Representations, {ICLR} 2023, Kigali, Rwanda, May 1-5, 2023}. OpenReview.net, 2023.
\newblock URL \url{https://openreview.net/pdf?id=XVjTT1nw5z}.

\bibitem[Lopez{-}Paz \& Oquab(2017)Lopez{-}Paz and Oquab]{DBLP:conf/iclr/Lopez-PazO17}
David Lopez{-}Paz and Maxime Oquab.
\newblock Revisiting classifier two-sample tests.
\newblock In \emph{5th International Conference on Learning Representations, {ICLR} 2017, Toulon, France, April 24-26, 2017, Conference Track Proceedings}. OpenReview.net, 2017.
\newblock URL \url{https://openreview.net/forum?id=SJkXfE5xx}.

\bibitem[Lueckmann et~al.(2021)Lueckmann, Boelts, Greenberg, Gon{\c{c}}alves, and Macke]{DBLP:conf/aistats/LueckmannBGGM21}
Jan{-}Matthis Lueckmann, Jan Boelts, David~S. Greenberg, Pedro~J. Gon{\c{c}}alves, and Jakob~H. Macke.
\newblock Benchmarking simulation-based inference.
\newblock In \emph{The 24th International Conference on Artificial Intelligence and Statistics, {AISTATS} 2021, April 13-15, 2021, Virtual Event}, volume 130 of \emph{Proceedings of Machine Learning Research}, pp.\  343--351. {PMLR}, 2021.
\newblock URL \url{http://proceedings.mlr.press/v130/lueckmann21a.html}.

\bibitem[Papamakarios et~al.(2017)Papamakarios, Murray, and Pavlakou]{DBLP:conf/nips/PapamakariosMP17}
George Papamakarios, Iain Murray, and Theo Pavlakou.
\newblock Masked autoregressive flow for density estimation.
\newblock In \emph{Advances in Neural Information Processing Systems 30}, pp.\  2338--2347, 2017.
\newblock URL \url{https://proceedings.neurips.cc/paper/2017/hash/6c1da886822c67822bcf3679d04369fa-Abstract.html}.

\bibitem[Papamakarios et~al.(2019)Papamakarios, Sterratt, and Murray]{DBLP:conf/aistats/PapamakariosS019}
George Papamakarios, David~C. Sterratt, and Iain Murray.
\newblock Sequential neural likelihood: Fast likelihood-free inference with autoregressive flows.
\newblock In Kamalika Chaudhuri and Masashi Sugiyama (eds.), \emph{The 22nd International Conference on Artificial Intelligence and Statistics, {AISTATS}}, volume~89 of \emph{Proceedings of Machine Learning Research}, pp.\  837--848. {PMLR}, 2019.
\newblock URL \url{http://proceedings.mlr.press/v89/papamakarios19a.html}.

\bibitem[Phan et~al.(2019)Phan, Pradhan, and Jankowiak]{phan2019composable}
Du~Phan, Neeraj Pradhan, and Martin Jankowiak.
\newblock Composable effects for flexible and accelerated probabilistic programming in numpyro.
\newblock \emph{arXiv preprint arXiv:1912.11554}, 2019.

\bibitem[Poh et~al.(2022)Poh, Samudre, {\'C}iprijanovi{\'c}, Nord, Khullar, Tanoglidis, and Frieman]{poh2022strong}
Jason Poh, Ashwin Samudre, Aleksandra {\'C}iprijanovi{\'c}, Brian Nord, Gourav Khullar, Dimitrios Tanoglidis, and Joshua~A Frieman.
\newblock Strong lensing parameter estimation on ground-based imaging data using simulation-based inference.
\newblock \emph{arXiv preprint arXiv:2211.05836}, 2022.

\bibitem[Pooladian et~al.(2023)Pooladian, Ben{-}Hamu, Domingo{-}Enrich, Amos, Lipman, and Chen]{DBLP:conf/icml/PooladianBDALC23}
Aram{-}Alexandre Pooladian, Heli Ben{-}Hamu, Carles Domingo{-}Enrich, Brandon Amos, Yaron Lipman, and Ricky T.~Q. Chen.
\newblock Multisample flow matching: Straightening flows with minibatch couplings.
\newblock In Andreas Krause, Emma Brunskill, Kyunghyun Cho, Barbara Engelhardt, Sivan Sabato, and Jonathan Scarlett (eds.), \emph{International Conference on Machine Learning, {ICML} 2023, 23-29 July 2023, Honolulu, Hawaii, {USA}}, volume 202 of \emph{Proceedings of Machine Learning Research}, pp.\  28100--28127. {PMLR}, 2023.
\newblock URL \url{https://proceedings.mlr.press/v202/pooladian23a.html}.

\bibitem[Rezende \& Mohamed(2015)Rezende and Mohamed]{DBLP:conf/icml/RezendeM15}
Danilo~Jimenez Rezende and Shakir Mohamed.
\newblock Variational inference with normalizing flows.
\newblock In \emph{Proceedings of the 32nd International Conference on Machine Learning, {ICML} 2015, Lille, France, 6-11 July 2015}, volume~37 of \emph{{JMLR} Workshop and Conference Proceedings}, pp.\  1530--1538. JMLR.org, 2015.
\newblock URL \url{http://proceedings.mlr.press/v37/rezende15.html}.

\bibitem[Saharia et~al.(2022)Saharia, Chan, Saxena, Li, Whang, Denton, Ghasemipour, Lopes, Ayan, Salimans, Ho, Fleet, and Norouzi]{DBLP:conf/nips/SahariaCSLWDGLA22}
Chitwan Saharia, William Chan, Saurabh Saxena, Lala Li, Jay Whang, Emily~L. Denton, Seyed Kamyar~Seyed Ghasemipour, Raphael~Gontijo Lopes, Burcu~Karagol Ayan, Tim Salimans, Jonathan Ho, David~J. Fleet, and Mohammad Norouzi.
\newblock Photorealistic text-to-image diffusion models with deep language understanding.
\newblock In \emph{Advances in Neural Information Processing Systems 35}, 2022.
\newblock URL \url{http://papers.nips.cc/paper\_files/paper/2022/hash/ec795aeadae0b7d230fa35cbaf04c041-Abstract-Conference.html}.

\bibitem[Schuldt et~al.(2021)Schuldt, Suyu, Meinhardt, Leal-Taix{\'e}, Ca{\~n}ameras, Taubenberger, and Halkola]{schuldt2021holismokes}
S~Schuldt, SH~Suyu, T~Meinhardt, L~Leal-Taix{\'e}, R~Ca{\~n}ameras, S~Taubenberger, and A~Halkola.
\newblock Holismokes-iv. efficient mass modeling of strong lenses through deep learning.
\newblock \emph{Astronomy \& Astrophysics}, 646:\penalty0 A126, 2021.

\bibitem[Sharrock et~al.(2022)Sharrock, Simons, Liu, and Beaumont]{DBLP:journals/corr/sharrock2022}
Louis Sharrock, Jack Simons, Song Liu, and Mark Beaumont.
\newblock Sequential neural score estimation: Likelihood-free inference with conditional score based diffusion models.
\newblock \emph{CoRR}, abs/2210.04872, 2022.
\newblock \doi{10.48550/ARXIV.2210.04872}.
\newblock URL \url{https://doi.org/10.48550/arXiv.2210.04872}.

\bibitem[Song et~al.(2021)Song, Sohl{-}Dickstein, Kingma, Kumar, Ermon, and Poole]{DBLP:conf/iclr/songscore21}
Yang Song, Jascha Sohl{-}Dickstein, Diederik~P. Kingma, Abhishek Kumar, Stefano Ermon, and Ben Poole.
\newblock Score-based generative modeling through stochastic differential equations.
\newblock In \emph{9th International Conference on Learning Representations, {ICLR} 2021, Virtual Event, Austria, May 3-7, 2021}. OpenReview.net, 2021.
\newblock URL \url{https://openreview.net/forum?id=PxTIG12RRHS}.

\bibitem[Talts et~al.(2018)Talts, Betancourt, Simpson, Vehtari, and Gelman]{talts2018validating}
Sean Talts, Michael Betancourt, Daniel Simpson, Aki Vehtari, and Andrew Gelman.
\newblock Validating bayesian inference algorithms with simulation-based calibration.
\newblock \emph{arXiv preprint arXiv:1804.06788}, 2018.

\bibitem[Tong et~al.(2023)Tong, Malkin, Huguet, Zhang, Rector{-}Brooks, Fatras, Wolf, and Bengio]{DBLP:journals/corr/tong2023}
Alexander Tong, Nikolay Malkin, Guillaume Huguet, Yanlei Zhang, Jarrid Rector{-}Brooks, Kilian Fatras, Guy Wolf, and Yoshua Bengio.
\newblock Conditional flow matching: Simulation-free dynamic optimal transport.
\newblock \emph{CoRR}, abs/2302.00482, 2023.
\newblock \doi{10.48550/ARXIV.2302.00482}.
\newblock URL \url{https://doi.org/10.48550/arXiv.2302.00482}.

\bibitem[Vegetti et~al.(2023)Vegetti, Birrer, Despali, Fassnacht, Gilman, Hezaveh, Levasseur, McKean, Powell, O'Riordan, et~al.]{vegetti2023strong}
S~Vegetti, S~Birrer, G~Despali, CD~Fassnacht, D~Gilman, Y~Hezaveh, L~Perreault Levasseur, JP~McKean, DM~Powell, CM~O'Riordan, et~al.
\newblock Strong gravitational lensing as a probe of dark matter.
\newblock \emph{arXiv preprint arXiv:2306.11781}, 2023.

\bibitem[Vegetti \& Koopmans(2009)Vegetti and Koopmans]{vegetti2009bayesian}
Simona Vegetti and L{\'e}on~VE Koopmans.
\newblock Bayesian strong gravitational-lens modelling on adaptive grids: objective detection of mass substructure in galaxies.
\newblock \emph{Monthly Notices of the Royal Astronomical Society}, 392\penalty0 (3):\penalty0 945--963, 2009.

\bibitem[Wildberger et~al.(2023)Wildberger, Dax, Buchholz, Green, Macke, and Sch{\"{o}}lkopf]{DBLP:conf/nips/WildbergerDBGMS23}
Jonas Wildberger, Maximilian Dax, Simon Buchholz, Stephen~R. Green, Jakob~H. Macke, and Bernhard Sch{\"{o}}lkopf.
\newblock Flow matching for scalable simulation-based inference.
\newblock In \emph{Advances in Neural Information Processing Systems 36}, 2023.
\newblock URL \url{http://papers.nips.cc/paper\_files/paper/2023/hash/3663ae53ec078860bb0b9c6606e092a0-Abstract-Conference.html}.

\bibitem[Wu et~al.(2023)Wu, Trippe, Naesseth, Blei, and Cunningham]{DBLP:conf/nips/WuTNBC23}
Luhuan Wu, Brian~L. Trippe, Christian~A. Naesseth, David~M. Blei, and John~P. Cunningham.
\newblock Practical and asymptotically exact conditional sampling in diffusion models.
\newblock In \emph{Advances in Neural Information Processing Systems 36}, 2023.
\newblock URL \url{http://papers.nips.cc/paper\_files/paper/2023/hash/63e8bc7bbf1cfea36d1d1b6538aecce5-Abstract-Conference.html}.

\end{thebibliography}
\bibliographystyle{iclr2025_conference}

\clearpage 

\appendix
\section*{{\LARGE Appendix}}
\section{Algorithms} \label{app:algorithms}

We include algorithms for training using flow matching and control signals, see algorithm \ref{algorithm_control}. For flow matching with self-conditioning, see algorithm \ref{algorithm_self_conditioning}.

\begin{figure}[h]
\begin{minipage}{0.49\textwidth}
\begin{algorithm}[H]
    \centering
    \caption{FM with Control Signals}\label{algorithm_control}
    \begin{algorithmic}
        \State \textbf{Input:} Training distribution $q_1$, pretrained network $v_\phi$, control network $v_\phi^C$, $\sigma_\mathrm{min}$
        \While{Training}
            \State $(\boldsymbol{\theta}_1, \boldsymbol{x}_o) \sim q_1; \; z \gets \mathcal{N}(0,I)$ 
            \State $\boldsymbol{\theta} \gets t \boldsymbol{\theta}_1 + (1-t) \boldsymbol{z}$
            \State \textcolor{blue}{$\boldsymbol{v} \gets \mathrm{stopgrad}(v_\phi(t,\boldsymbol{\theta}, \boldsymbol{x}_o))$}
            \State \textcolor{blue}{$\hat{\boldsymbol{\theta}}_1 \gets \boldsymbol{\theta} + (1-t)\boldsymbol{v}$}
            \State \textcolor{blue}{$\boldsymbol{c} \gets \mathrm{control}(\hat{\boldsymbol{\theta}}_1,  \boldsymbol{x}_o)$}
            \State \textcolor{blue}{$\Tilde{\boldsymbol{v}} \gets v_\phi^C(t, \boldsymbol{v}, \boldsymbol{c}) + \boldsymbol{v}$}
            \State $u_t(\boldsymbol{\theta}|\boldsymbol{\theta}_1,  \boldsymbol{x}_o) \gets \tfrac{\boldsymbol{\theta}_1 - (1-\sigma_{\mathrm{min}})\boldsymbol{\theta}}{1 - (1-\sigma_\mathrm{min})t}$
            \State $\mathcal{L}_{\mathrm{CFM}} \gets \norm{\Tilde{\boldsymbol{v}} - u_t(\boldsymbol{\theta}|\boldsymbol{\theta}_1,  \boldsymbol{x}_o)}_2^2$
            \State $\theta \gets \mathrm{Update}(\phi, \nabla _\phi \mathcal{L}_\mathrm{CFM}(\phi))$
        \EndWhile
        \State \textbf{return: } $v_\phi$, $v_\phi^C$
    \end{algorithmic}
\end{algorithm}
\end{minipage}
\hfill 
\begin{minipage}{0.5\textwidth}
\begin{algorithm}[H]
    \centering
    \caption{FM with Self-conditioning}\label{algorithm_self_conditioning}
    \begin{algorithmic}
         \State \textbf{Input:} Training distribution $q_1$, flow network $v_\phi$, $\sigma_\mathrm{min}$
        \While{Training}
            \State $(\boldsymbol{\theta}_1,  \boldsymbol{x}_o) \sim q_1;\, z \gets \mathcal{N}(0,I);\, s \gets \mathcal{U}(0,1)$
            \State $\boldsymbol{\theta} \gets t \boldsymbol{\theta}_1 + (1-t) \boldsymbol{z}; \; \hat{\boldsymbol{\theta}}_1 \gets 0$
            \State \textcolor{blue}{$\boldsymbol{v} \gets \mathrm{stopgrad}(v_\theta(t,[\boldsymbol{\theta}, \hat{\boldsymbol{\theta}}_1], \boldsymbol{x}_o))$}
            \textcolor{blue}{\If{$s > 0.5$}}
                \State $\quad$ \textcolor{blue}{$\hat{\boldsymbol{\theta}}_1 \gets \boldsymbol{\theta} + (1-t) \boldsymbol{v}$}
            \State \textcolor{blue}{$\Tilde{\boldsymbol{v}} \gets v_\phi(t, [\boldsymbol{\theta}, \hat{\boldsymbol{\theta}}_1],  \boldsymbol{x}_o)$}
            \State $u_t(\boldsymbol{\theta}|\boldsymbol{\theta}_1,  \boldsymbol{x}_o) \gets \tfrac{\boldsymbol{\theta}_1 - (1-\sigma_{\mathrm{min}})\boldsymbol{\theta}}{1 - (1-\sigma_\mathrm{min})t}$
            \State $\mathcal{L}_{\mathrm{CFM}} \gets \norm{\Tilde{\boldsymbol{v}} - u_t(\boldsymbol{\theta}|\boldsymbol{\theta}_1, \boldsymbol{x}_o)}_2^2$
            \State $\theta \gets \mathrm{Update}(\phi, \nabla _\phi \mathcal{L}_\mathrm{CFM}(\phi))$
        \EndWhile
        \State \textbf{return: } $v_\phi$
    \end{algorithmic}
\end{algorithm}
\end{minipage}
\end{figure}

\clearpage

\section{Simulation-based Inference} \label{sec:sbi}

\paragraph{Baselines comparison in section \ref{sec:sbi_tasks}}

For a fairer comparison, we set up all baseline methods with a similar number of network weights and available compute time. 

We train all baselines and flow matching with a batch size of 512 on the largest $10^5$ simulation budges for all tasks. For optimization, we apply Adam \citep{DBLP:journals/corr/KingmaB14} with default settings and constant learning rate of $10^{-4}$ and weight decay $2 \times 10^{-5}$. 

All network architectures are chosen to have a similar number of ca. $3 \cdot 10^5$ parameters. For flow matching and continuous normalizing flows (CNFs), we use the same architecture based on a dense feed-forward neural net with skip connections using 8 residual blocks with each 128 neurons and \emph{elu} activation. As input, we concatenate time $t$ and $\boldsymbol{\theta}_t$.
For Neural Spline Flow \citep{DBLP:conf/nips/DurkanB0P19} and FFJORD \citep{DBLP:conf/iclr/GrathwohlCBSD19}, we adopt the released implementation by the authors.

Depending on the time per epoch for each method, we modify the number of epochs and steps per epoch to allow all methods to train for a similar amount of time, ensuring a sufficient window for convergence. For NSF, we train for 1 000 epochs, for flow matching for 2 000 epochs, and for FFJORD and CNF 100 epochs.  

\paragraph{Flow matching with optimized hyperparameters} For the experiments in section \ref{sec:training_variants} and section \ref{sec:simulator_feedback}, we adopt the hyperparameters and network architecture from \citet{DBLP:conf/nips/WildbergerDBGMS23}, which is based on a hyperparamter grid search. The hyperparameters for each task are listed in table \ref{tab:sbi_hyperparameters}. Otherwise, we follow the implementation as provided by the authors. 

\begin{table}[h]
    \centering
    \caption{Hyperparameters for SBI from \cite{DBLP:conf/nips/WildbergerDBGMS23}.}
    \label{tab:sbi_hyperparameters}
    \begin{tabular}{lcccl}
       \bf Task & \bf Time $\alpha$ & \bf Batch size & \bf Learning rate & \bf Residual blocks \\ 
       \hline \\
       LV & 1 & 32 & $10^{-3}$ & [32, 64, 128, 256, 5$\times$512, 256, 128, 64, 32]  \\ 
       SLCP & -0.5 & 256 & $5\cdot10^{-4}$ & [32, 64, 128, 256, 5$\times$512, 256, 128, 64, 32] \\
       SIR & 4 & 256 & $5\cdot 10^{-4}$ & [32, 64, 128, 256, 7$\times$512, 256, 128, 64, 32] \\
       TM & 4 & 64 & $2\cdot 10^{-4}$ & [32, 64, 128, 256, 512, 3$\times$1024, 512, 128, 64, 32]\\
    \end{tabular}
\end{table}

\begin{figure}[h]
    \centering
    \begin{subfigure}[]{0.4\linewidth}
        \includegraphics[height=.7\linewidth]{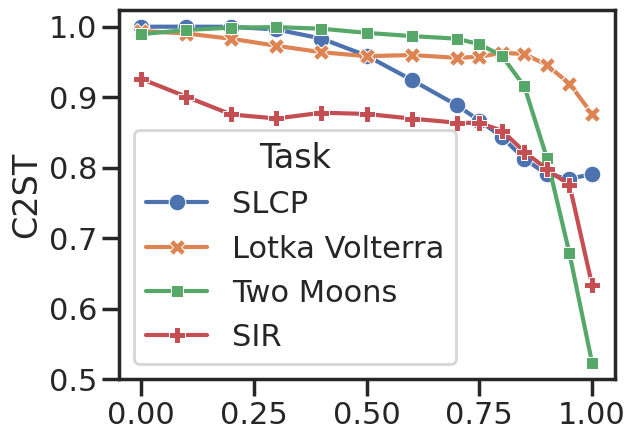}
        \caption{}
    \end{subfigure}
    \begin{subfigure}[]{0.4\linewidth}
        \includegraphics[height=.7\linewidth]{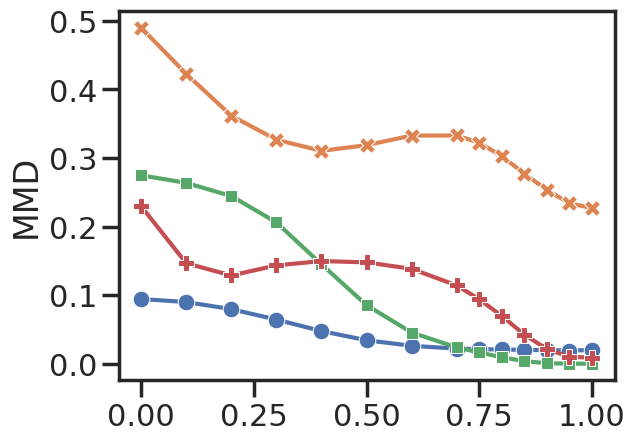}
        \caption{}
    \end{subfigure}
    \caption{(a) and (b): C2ST score and MMD for predictive samples \smash{$\hat{\boldsymbol{\theta}}_1$}. The $x$-axis shows from which we compute the predictive sample.}
    \label{fig:sbi_1_step}
\end{figure}

\paragraph{Analyzing the 1-step estimate} We simulate the flow ODE from the sampling distribution at $t=0$ until $t^*$ ($x$-axis). Then, we compute the posterior in a single step by linearly extrapolating the flow, see \eqref{eq:1_step_prediction}, to obtain the estimate \smash{$\hat{\boldsymbol{\theta}}_1$}. Results are shown in \figref{fig:sbi_1_step}.

\clearpage

\begin{figure}[h]
    \centering
    \begin{subfigure}[]{0.4\linewidth}
        \includegraphics[height=.7\linewidth]{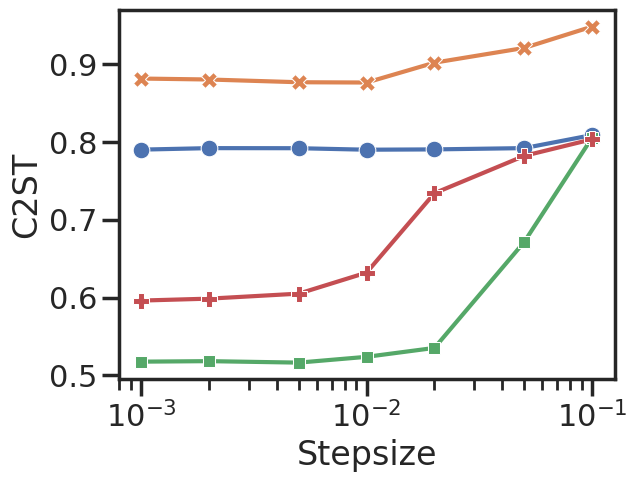}
        \caption{}
    \end{subfigure}
    \begin{subfigure}[]{0.4\linewidth}
        \includegraphics[height=.7\linewidth]{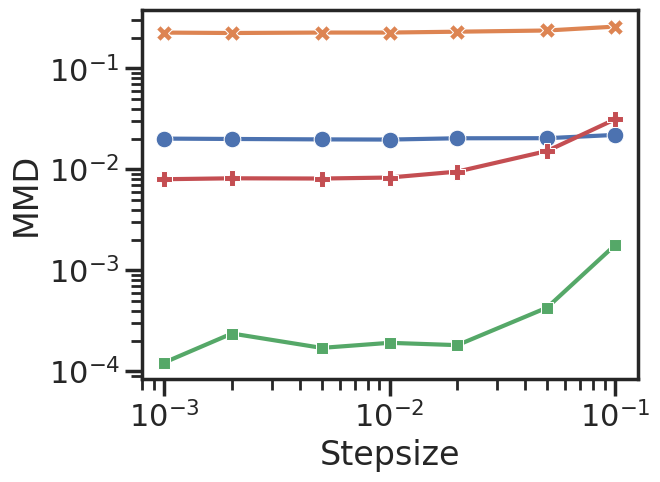}
        \caption{}
    \end{subfigure}
    \caption{(a) and (b): C2ST score and MMD vs. step size during inference.}
    \label{fig:sbi_stepsize}
\end{figure}

\paragraph{Analyzing step size} We analyze the influence of the step size of the ODE solver on the quality of the posterior distribution as shown in \figref{fig:sbi_stepsize}.

\begin{figure}[h]
\centering
\includegraphics[width=.6\textwidth]{pictures/iclr/legend.pdf}
 \\
\includegraphics[width=.36\textwidth]{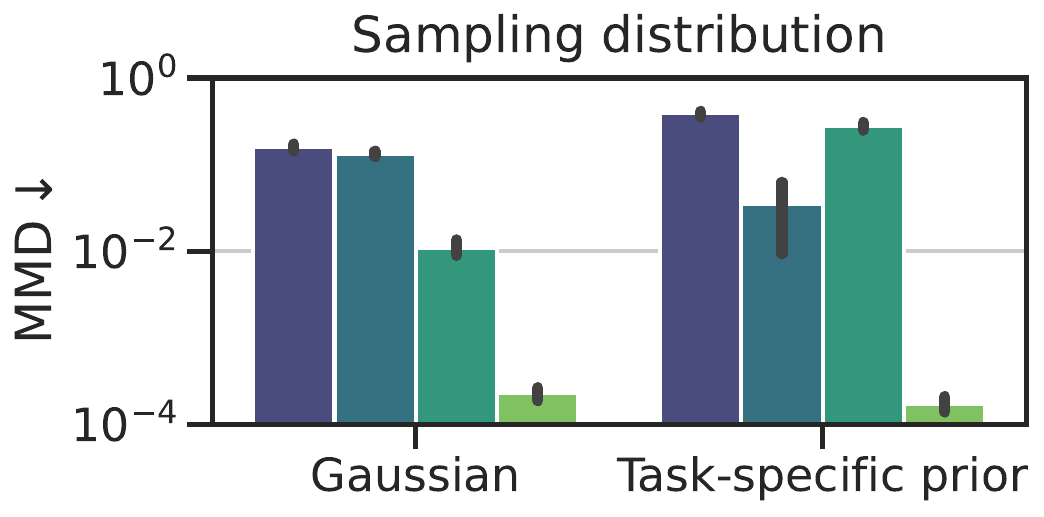}
\includegraphics[width=.30\textwidth]{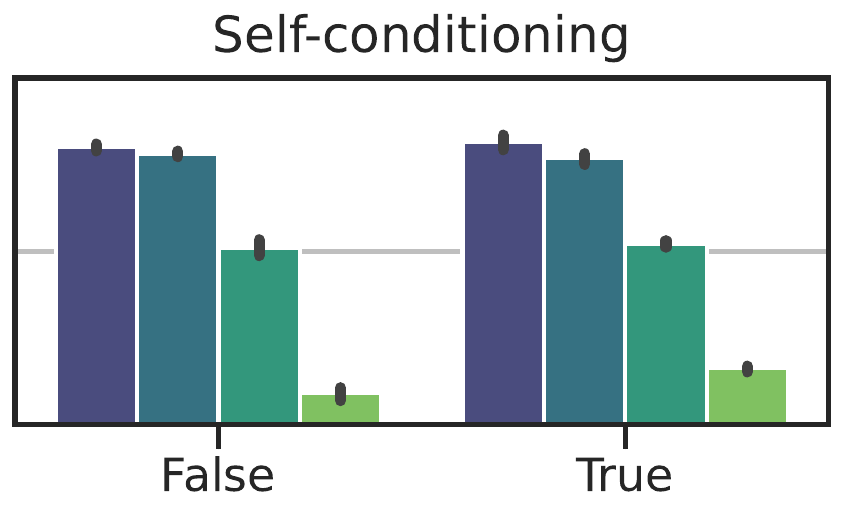}
\includegraphics[width=.30\textwidth]{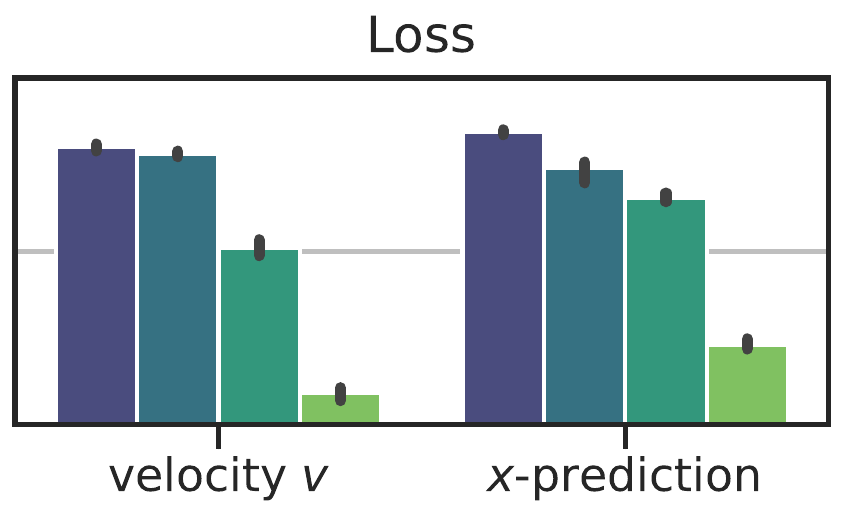}
\caption{
Evaluation of SBI tasks using different variants of flow matching training. Lower MMD scores are better.}
\label{fig:sbi_benchmarks_mmd}
\end{figure}

\paragraph{Additional results for maximum mean discrepancy} For the evaluation in section \ref{sec:training_variants}, we show additional results for the maximum mean discrepancy \citep[MMD]{gretton2012kernel} in \figref{fig:sbi_benchmarks_mmd}.

\clearpage

\section{Strong Gravitational Lensing}
\label{app:strong_lensing}
We consider the following models:
\begin{itemize}
    \item For modeling the lens we use an SIE model with 6 parameters: the Einstein radius $\theta_E$, the ellipticities $e_1$ and $e_2$ and $x_\mathrm{center}$ and $y_\mathrm{center}$. There is  shear, for which we only consider  $\gamma_1$ and $\gamma_2$ as free parameters.
    \item The source is modeled by a Sersic profile with free parameters being the amplitude, the half-light radius, the Sersic index $n$, the ellipticities $e_1$ and $e_2$ as well as the positions $x_\mathrm{center}$ and $y_\mathrm{center}$.
    \item The lens light is modeled in the same way as the source, although when generating the mock  data, we fix the position as well as ellipticities to be the same as the lens mass model. For training and inference, we infer positions for both lens mass and lens light model, so the model could produce different values for them. The MCMC methods use the same parameter for both lens light and lens mass position.  
\end{itemize}
We list all priors in table \ref{tab:prior_lens_mass}, table \ref{tab:prior_source_light} and table \ref{tab:prior_lens_light}. We do not have priors on the ellipticities $e_1$ and $e_2$ directly, but we obtain them from priors on the position angle and axis ratio. Also, we obtain the shear parameters from $\gamma_1$ and $\gamma_2$ from $\phi_\mathrm{ext}$ and $\gamma_\mathrm{ext}$ by converting them polar to cartesian coordinates. 
For SBI, we also include the two parameters $\mathrm{ra}_0$ and $\mathrm{dec}_0$ for the shear, which are always set to 0 when generating the training data sets, but in general our network could infer other values. Overall, there are $23$ parameters for $v_\theta$, which fully describe the simulation setup. However, in our dataset there are only $17$ free parameters. The MCMC methods only infer the reduced set of parameters, making use of the dependencies between them. 

\paragraph{Measurement instruments}

Observations have $160$ times $160$ pixels. The pixel size is $0.04$ arc seconds. We use a Gaussian points spread function (PSF) with full width at half maximum (FWHM) of 0.3.
The there is Gaussian background noise with a root mean-squared values of $0.01$ and an exposure time of 1000s.

\paragraph{Setup of MCMC-based methods}
We setup both baselines methods as follows: 
\begin{enumerate}
    \item Hamiltonian Monte Carlo: we use the No-U-Turn samples with a maximum tree depth of $10$ and 5 000 warmup steps. 
    \item Affine-Invariant Ensemble Sampling: we use DEMove and StretchMove both with probability $0.5$. There are $400$ chains and we warm up for 20 000 steps before starting sampling.
\end{enumerate}
Both methods are implemented in numpyro and optimized with JAX, so their runtimes are comparable with each other.

\paragraph{Network architectures and training}

\begin{itemize}
    \item Our flow network $v_\phi$ comprises a lightweight feature extraction network, representated by a CNN, which is consists of $6$ downsampling blocks with 1 layer each a $32$ channels and kernel size 3. As postprocessing of the output, we apply GroupNorm, silu and an additional 2DConv layer with kernel size $3$ and a single channel. We reshape the output and feed it through a final dense layer. The output of the feature extraction has the same dimensionality as the parameters $\boldsymbol{\theta}$.
    \item An additional dense feed-forward neural network receives the concatenated the time $t$, $\boldsymbol{\theta}_t$ and extracted features as input. The feed-forward neural neural networks consists of 8 residual blocks with hidden dimension 128 and elu activation.
    \item The control network $v_\phi^C$ is represented by a small feed-forward neural network, consisting of 3 residual blocks with 64 hidden layers and 3 residual blocks with 32 hidden layers. We condition each block on the time via gated linear units and use a 16 dimensional time embedding. 
\end{itemize}
For training, we use a batch size of 256 for the flow network $v_\phi$. 
When training $v_\phi^C$, we decrease the batch size to 16. We use the Adam optimizer with a learning rate of $10^{-4}$ and weight decay of $10^{-5}$. Training $v_\phi$ was done on a single NVIDIA Ampere A100 GPU for ca. $45$ hours. We trained $v_\phi^C$ for an additional $24$ hours. A lot of the training time was spent on running evaluation metrics, so it can be substantially improved.  
\begin{table}[t]
\parbox{.45\linewidth}{
    \centering
    \caption{Priors for lens mass model parameters}
    \label{tab:prior_lens_mass}
    \begin{tabular}{l|l}
         Parameter & Prior \\
         \midrule
         $x_\mathrm{center}$ & $\mathcal{U}(-0.2, 0.2)$ \\
         $y_\mathrm{center}$ & $\mathcal{U}(-0.2, 0.2)$ \\
         position angle $\phi$ & $\mathcal{U}(0, 180)$ \\
         axis ratio $q$ & $\mathcal{U}(0.25, 1)$ \\
         external shear orientation $\phi_\mathrm{ext}$ & $\mathcal{U}(0, 180)$ \\
         external shear strength $\gamma_\mathrm{ext}$ & $\mathcal{U}(0, 0.1)$ \\
         Einstein radius $\theta_E$ & $\mathcal{U}(0.5, 2.0)$
    \end{tabular}
} \hfill
\parbox{.45\linewidth}{
    \centering
    \caption{Priors for the source light}
    \label{tab:prior_source_light}
    \begin{tabular}{l|l}
         Parameter & Prior \\
         \midrule
         amplitude & $\mathcal{U}(5.0, 10.0)$ \\
         half-light radius & $\mathcal{U}(0.5, 2.0)$ \\      Sersic index $n$ & 
         $\mathcal{U}(1.5, 4.0)$ \\ 
         position angle $\phi$ & $\mathcal{U}(0, 180)$ \\
         axis ratio $q$ & $\mathcal{U}(0.25, 1)$ \\
         $x_\mathrm{center}$ & $\mathcal{U}(-0.2, 0.2)$ \\
         $y_\mathrm{center}$ & $\mathcal{U}(-0.2, 0.2)$ \\
    \end{tabular}
}
\end{table}
\begin{table}[t]
    \centering
    \caption{Priors for the lens light}
    \label{tab:prior_lens_light}
    \begin{tabular}{l|l}
         Parameter & Prior \\
         \midrule
         amplitude & $\mathcal{U}(5.0, 10.0)$ \\
         half-light radius & $\mathcal{U}(0.5, 2.0)$ \\     Sersic index $n$ & 
         $\mathcal{U}(1.5, 4.0)$ \\ 
    \end{tabular}
\end{table}

\clearpage

\subsection{Diffusion Posterior Sampling (DPS)}\label{sec:dps}

We setup diffusion posterior sampling \cite{DBLP:conf/iclr/ChungKMKY23} as an additional baseline. The training dataset is the same as in \ref{sec:strong_gravitational_lensing}, however since the diffusion model is unconditional, we drop any conditioning information. 

\paragraph{Network architecture and training} The neural network architecture is a multilayer perceptron MLP with 8 residual blocks and 128 neurons each. The activation function is \emph{elu}. As optimizer, we use Adam with weight decay ($10^{-5}$). We train for $2000$ epochs and for each epoch we sample $1000$ batches from the dataset using a batch size of 4. We train the network as a denoising diffusion probabilistic model (DDPM) following \cite{DBLP:conf/nips/HoJA20}.

\paragraph{Unconditional generation} Below, in figure \ref{fig:dps_unconditional_generation}, we visualize three samples generated by unconditionally sampling from the model. The observations are created using the lensing simulation code with the generated samples as input.  

\begin{figure}[h]
    \centering
    \includegraphics[width=0.8\linewidth]{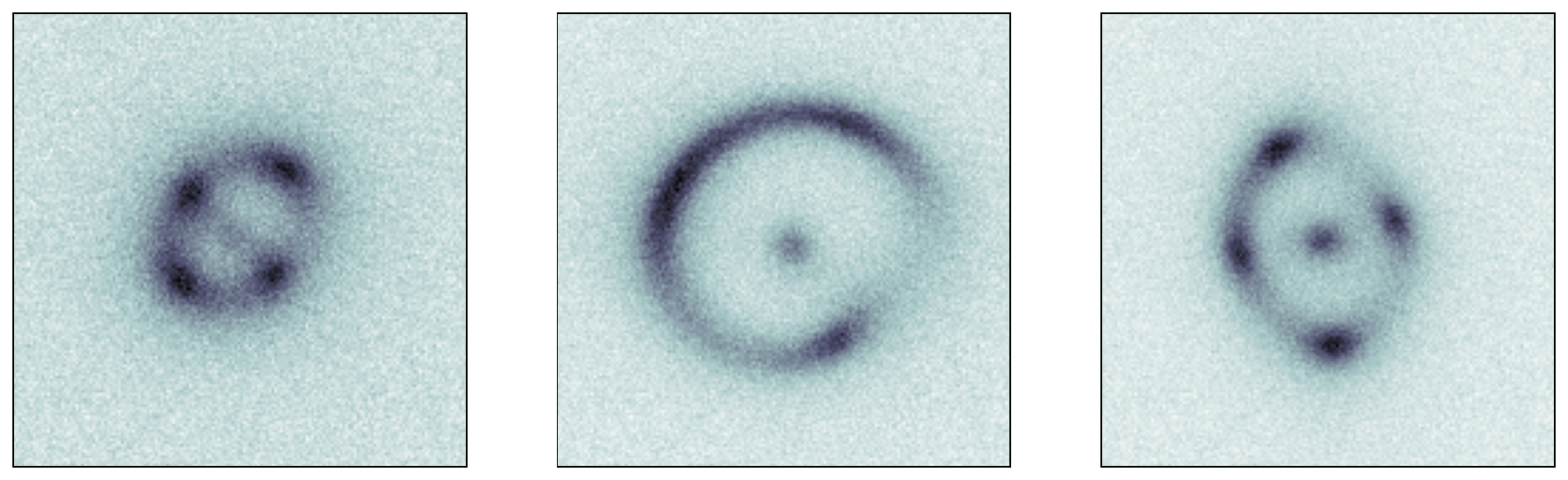}
    \caption{Visualization of unconditionally generated lensing systems.}
    \label{fig:dps_unconditional_generation}
\end{figure}

\paragraph{Inference} We directly follow \cite{DBLP:conf/iclr/ChungKMKY23} Algorithm 1 for inference, where the measurement operator $\mathcal{A}$ is replaced by the lensing simulation code. The step size in the algorithm is defined via a hyperparameter $\zeta$, which needs to be finetuned depending on the problem. We empirically test different values for $\zeta$ to find an optimal choice. Our results are shown in table \ref{tab:dps_eval}. In this evaluation, we only model a smaller number of systems ($n=25$).

\begin{table}[h]
    \centering
    \begin{tabular}{lcccccc}
         $\zeta$ & 0.0 & 0.0005 & 0.001 & 0.005 & 0.01 & 0.05 \\ \hline \\
         Avg. $\chi^2$ & 28.15 & 16.20 & \bf 9.98 & 10.07 & 12.98 & 12.64 \\
         Min. $\chi^2$ & 15.14 & 3.84 & 3.07 & 1.58 & \bf 1.40 & 1.53 \\
    \end{tabular}
    \caption{Evaluation of DPS and choosing $\zeta$.}
    \label{tab:dps_eval}
\end{table}

\clearpage

\subsection{Simulation-based Calibration} \label{sec:sbc}

\begin{figure}[h]
    \centering
    \begin{subfigure}[t]{0.3\linewidth}
        \includegraphics[width=\linewidth]{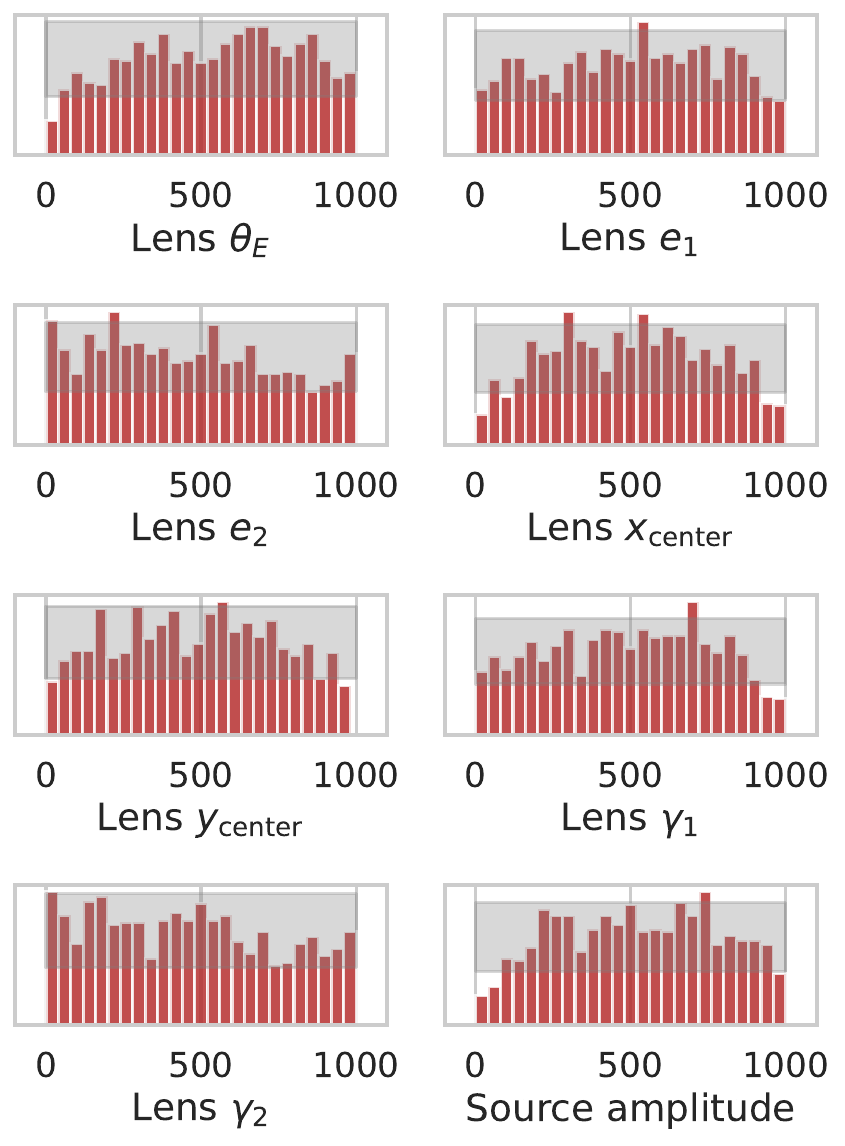}
        \caption{AIES}
    \end{subfigure}
    \hfill
    \begin{subfigure}[t]{0.3\linewidth}
        \includegraphics[width=\linewidth]{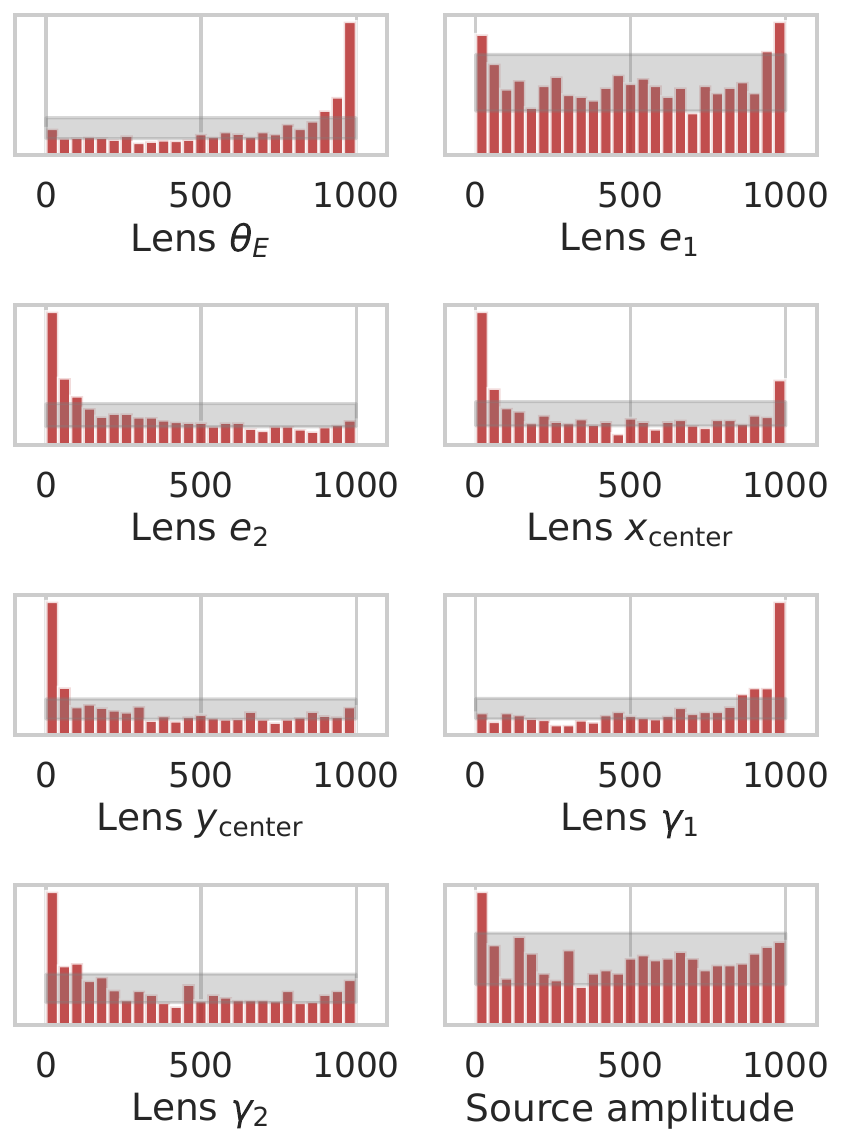}
        \caption{Flow Matching}
    \end{subfigure}
    \hfill
    \begin{subfigure}[t]{0.3\linewidth}
        \includegraphics[width=\linewidth]{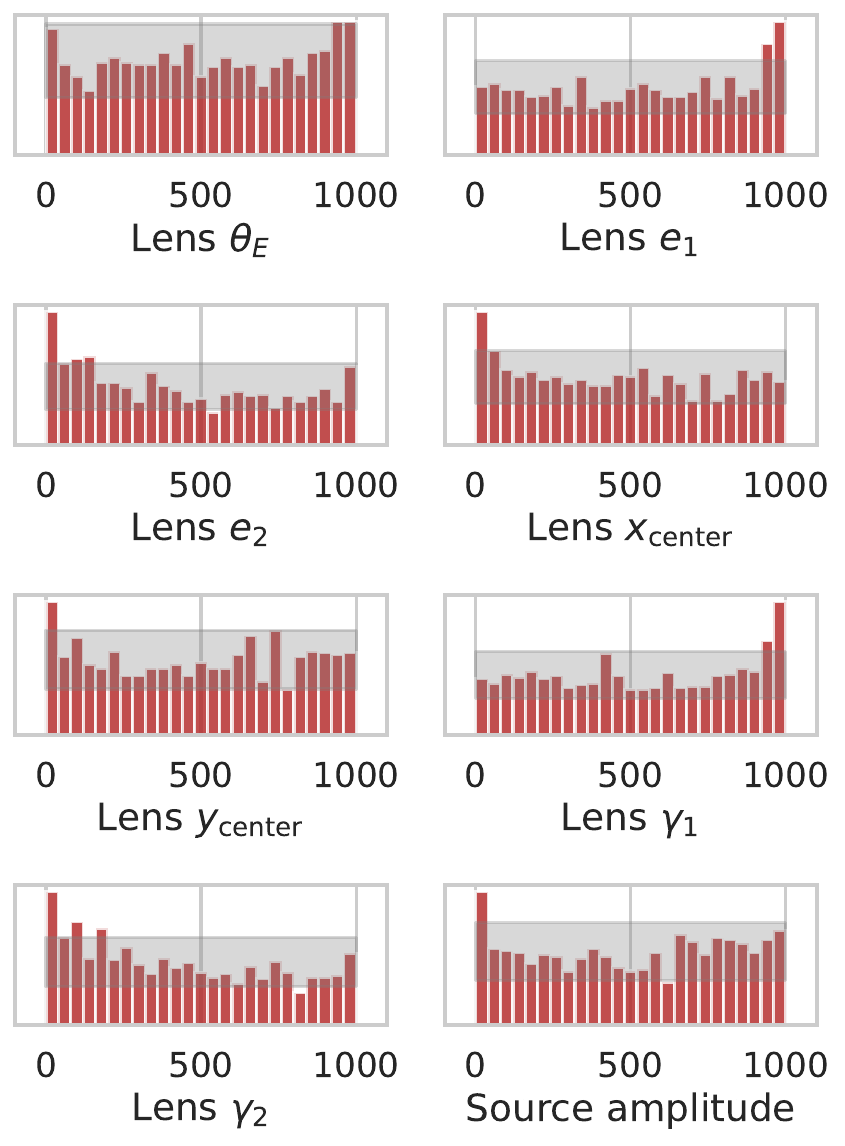}
        \caption{Flow Matching + Simulator}
    \end{subfigure}
    \caption{Simulation-based calibration histograms for different inference methods.}
    \label{fig:eval_sbc_lens}
\end{figure}

We use simulation-based calibration \cite[SBC]{talts2018validating} as an additional evaluation method to assess the correctness of the posterior distributions. We adopt the SBC implementation from the Python package \emph{sbi}.
Below, in \figref{fig:eval_sbc_lens}, we show histograms for 8 parameters based on $n=1000$ lens systems with $L=1000$ posterior samples each. 

\clearpage

\section{Posteriors and Reconstructions for Lens Modeling}

\begin{figure}
    \centering
    \includegraphics[width=\textwidth]{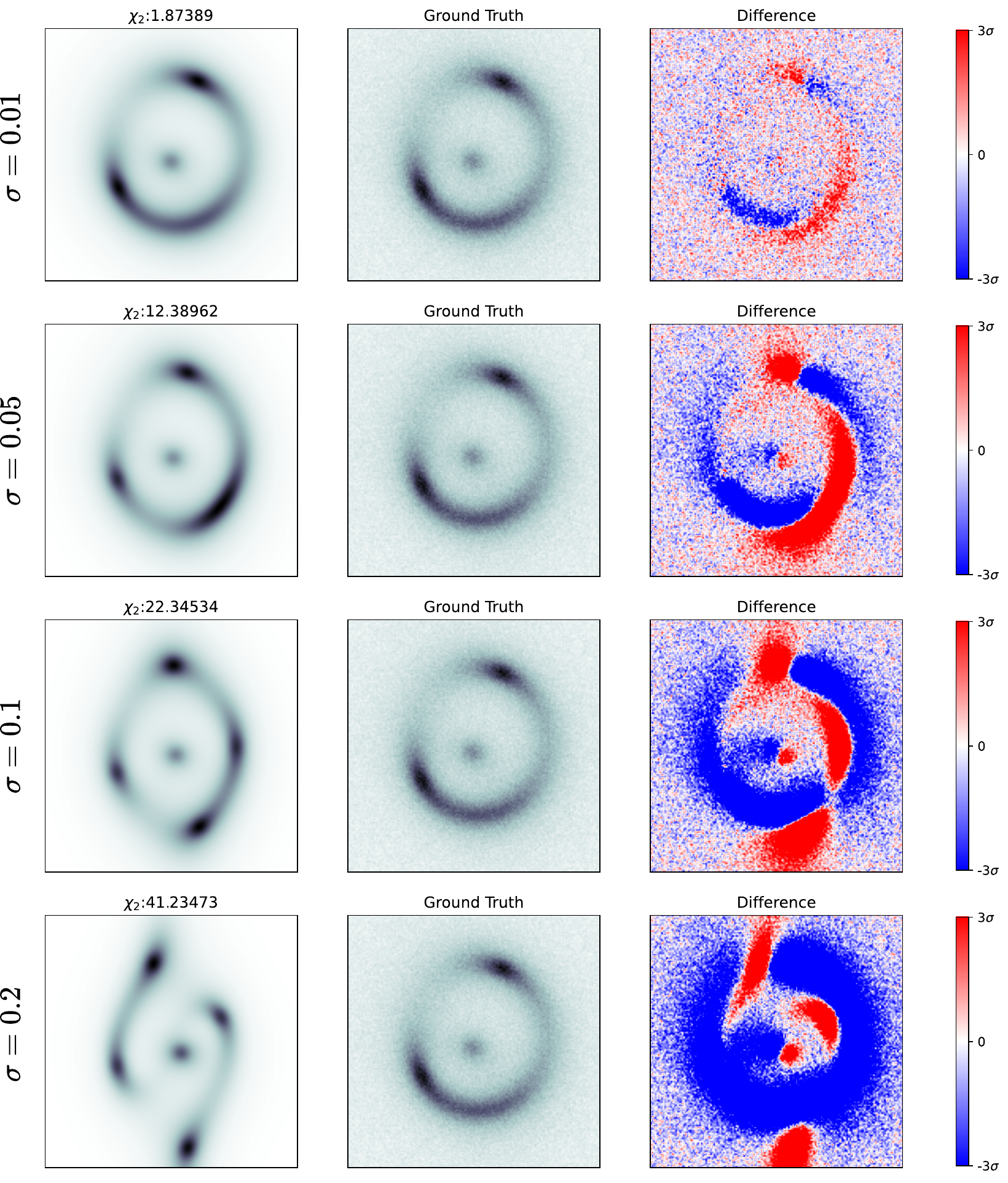}
    \caption{We show how a small noise $\sigma$ affects the simulated observation. We add a normal Gaussian with mean $0$ and standard deviation $\sigma$ to a lens system's ground truth parameters $x$. Then, we plot the simulated observation based on the noised parameters and show the residuals.}
    \label{fig:noise_reconstruction}
\end{figure}

\begin{figure}
    \centering
    \includegraphics[width=\textwidth]{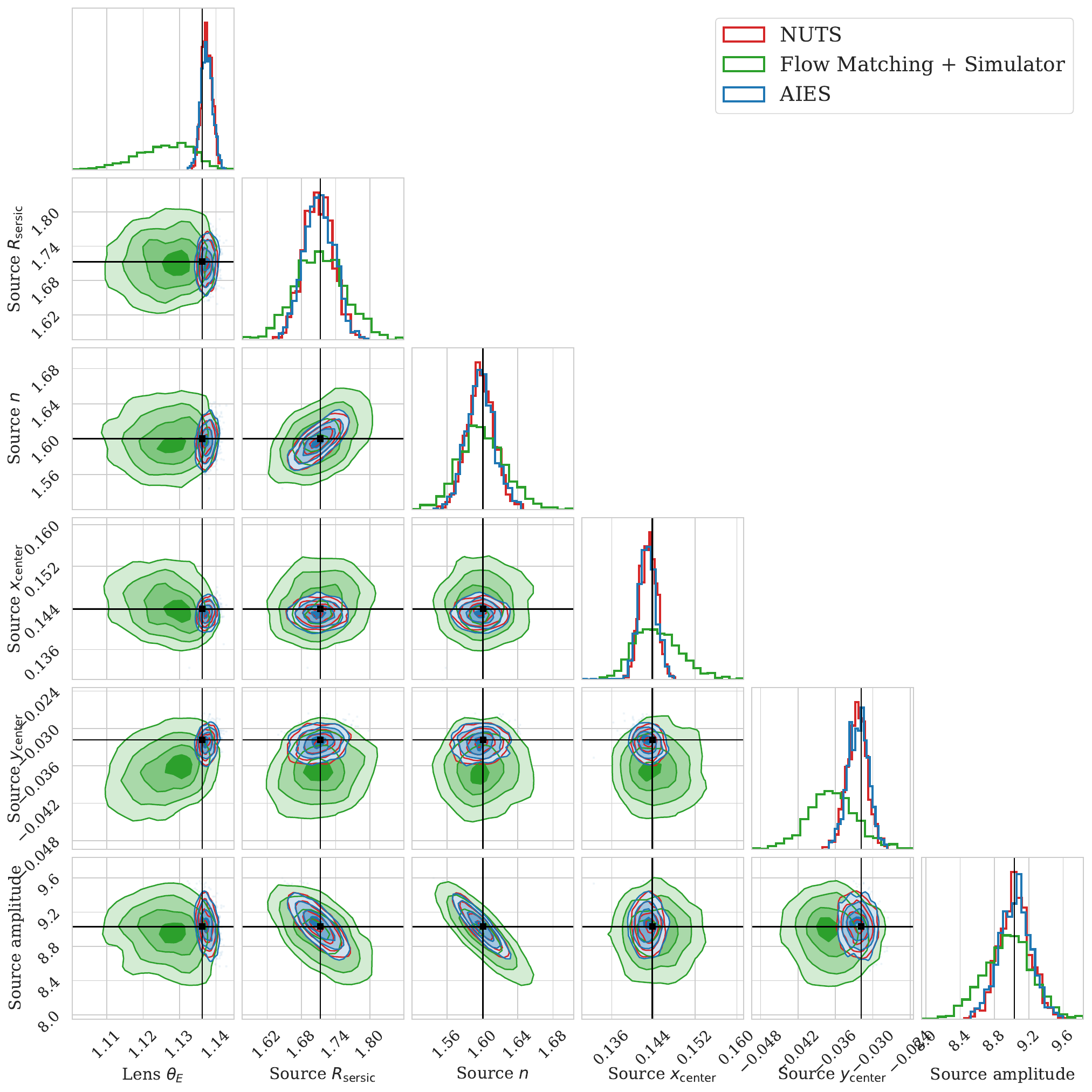}
    \caption{Posterior plot for system 1.}
    \label{fig:corner_lensing_system_1}
\end{figure}

\begin{figure}
    \centering
    \includegraphics[width=\textwidth]{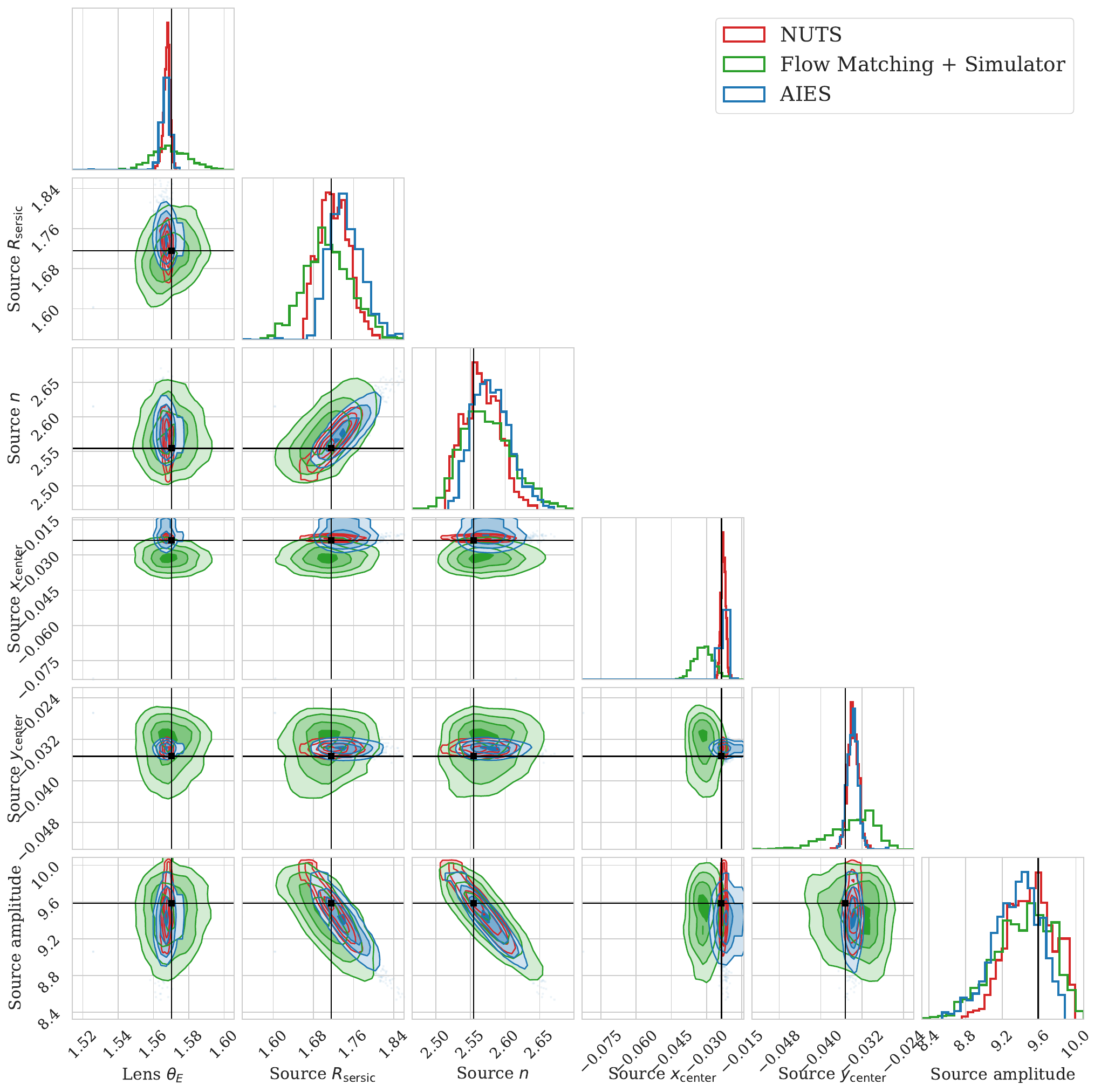}
    \caption{Posterior plot for system 6.}
    \label{fig:corner_lensing_system_2}
\end{figure}

\begin{figure}
    \centering
    \includegraphics[width=\textwidth]{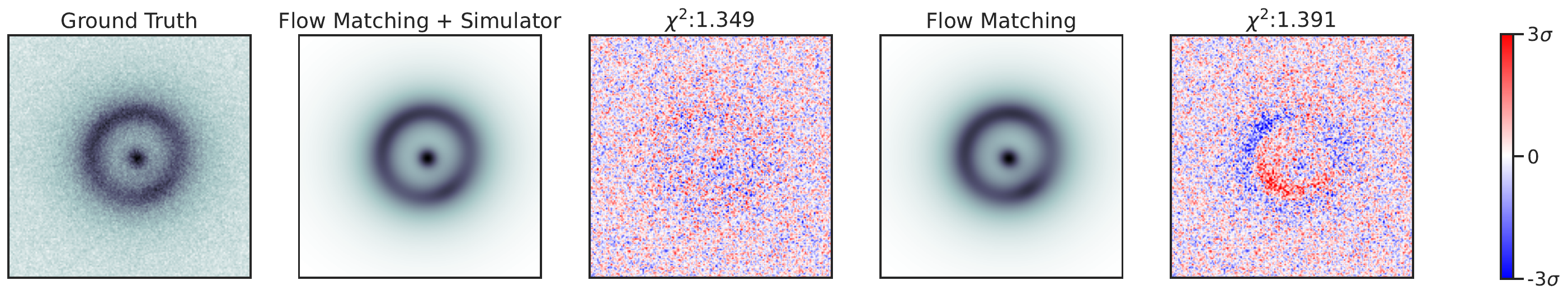}
    \includegraphics[width=\textwidth]{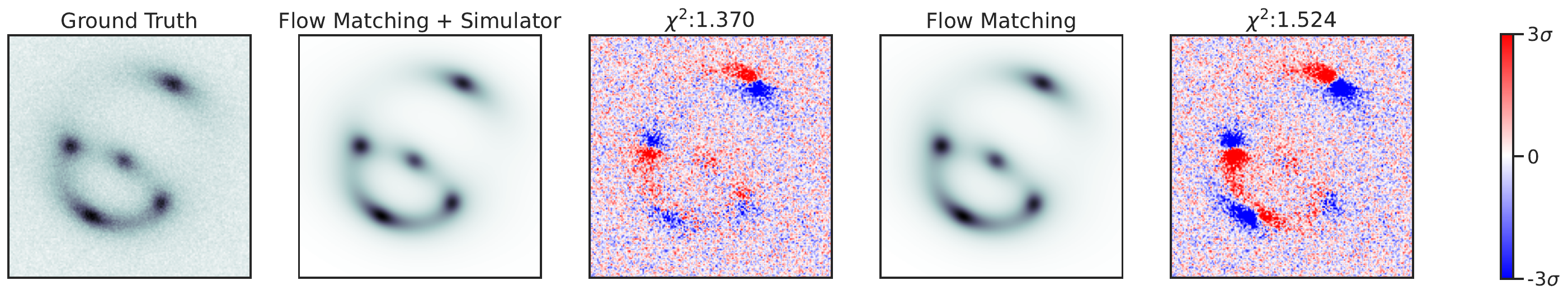}
    \includegraphics[width=\textwidth]{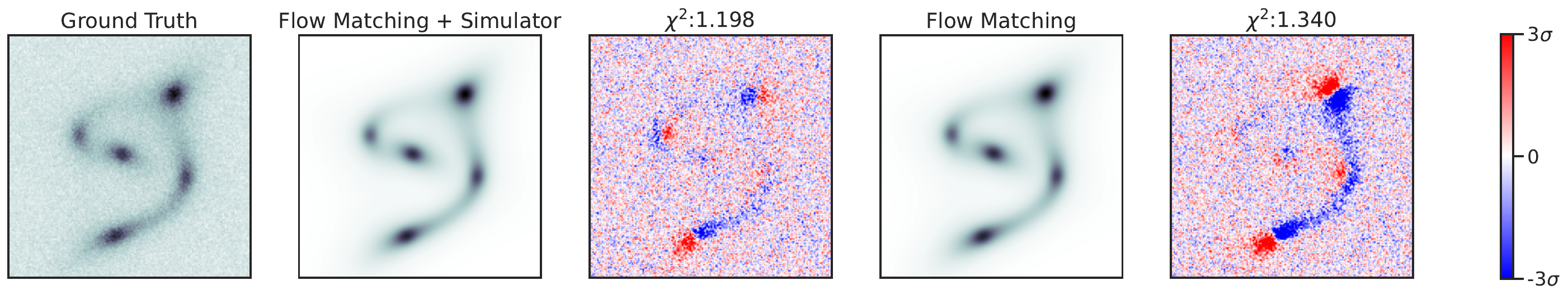}
    \includegraphics[width=\textwidth]{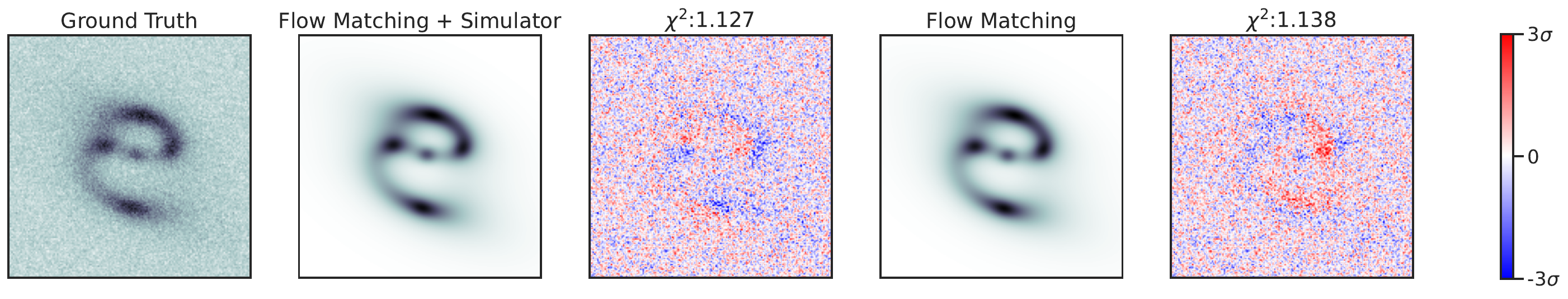}
    \includegraphics[width=\textwidth]{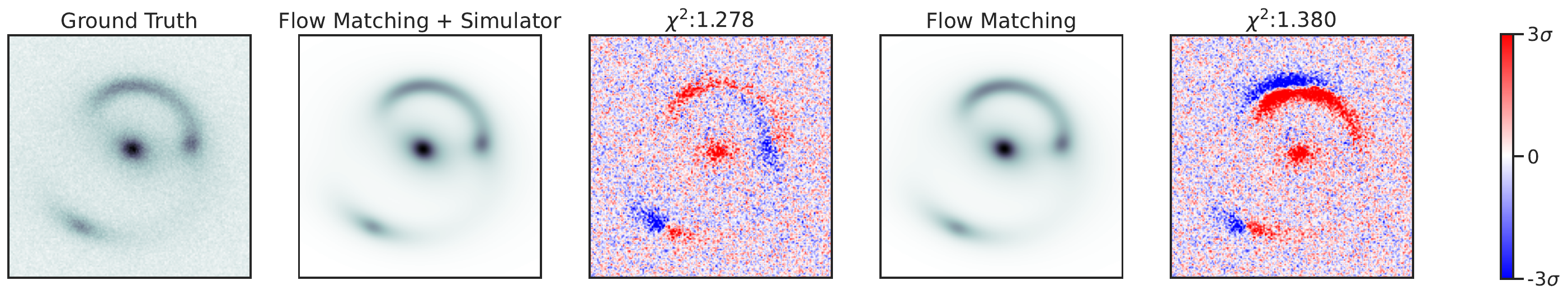}
    \includegraphics[width=\textwidth]{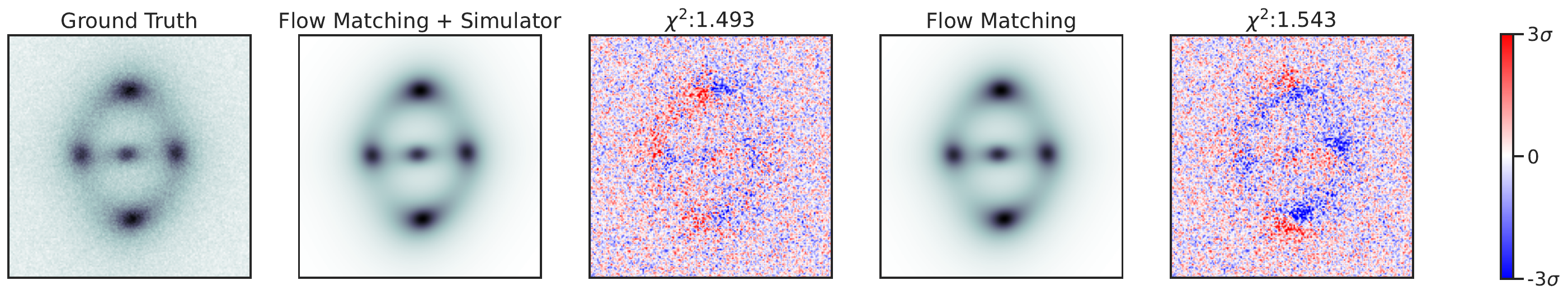}
    \caption{Modeling of different lens systems: system 1 (top) to system 6 (bottom).}
    \label{fig:lenses_modeling}
\end{figure}

We show how small perturbations in the lens system's parameters affect the simulated observation in figure \ref{fig:noise_reconstruction}.
We show extended plots of the posteriors in \figref{fig:corner_lensing_system_1} for lens system 1 and \figref{fig:corner_lensing_system_2} for lens system 6. 
Additionally, we show reconstructions based on flow matching with and without simulator feedback of lens systems 1 to 6 in \figref{fig:lenses_modeling}

\end{document}